\documentclass[runningheads]{llncs}

\usepackage{eccv}       

\usepackage{eccvabbrv}
\usepackage{xcolor}
\usepackage{algorithm}
\usepackage{pifont}
\usepackage{amssymb}           
\newcommand{\xmark}{\ding{55}} 
\usepackage{algpseudocode}
\usepackage{amssymb}
\usepackage{amsmath}
\usepackage{booktabs}
\usepackage{wrapfig}
\usepackage{graphicx} 
\usepackage{colortbl}
\usepackage{multirow}
\usepackage{hyperref}
\usepackage{makecell}
\usepackage{graphicx}
\usepackage{caption}
\usepackage{etoolbox}
\usepackage{graphicx}
\usepackage{graphicx}
\usepackage{booktabs}
\usepackage{graphicx}
\usepackage{caption}
\usepackage{xcolor}
\usepackage{bm}
\newcommand{\hlrow}{\rowcolor{black!6}}

\usepackage[accsupp]{axessibility}  

\usepackage{hyperref}

\usepackage{orcidlink}

\newcommand{\insertteaser}{
    \includegraphics[width=0.95\linewidth]{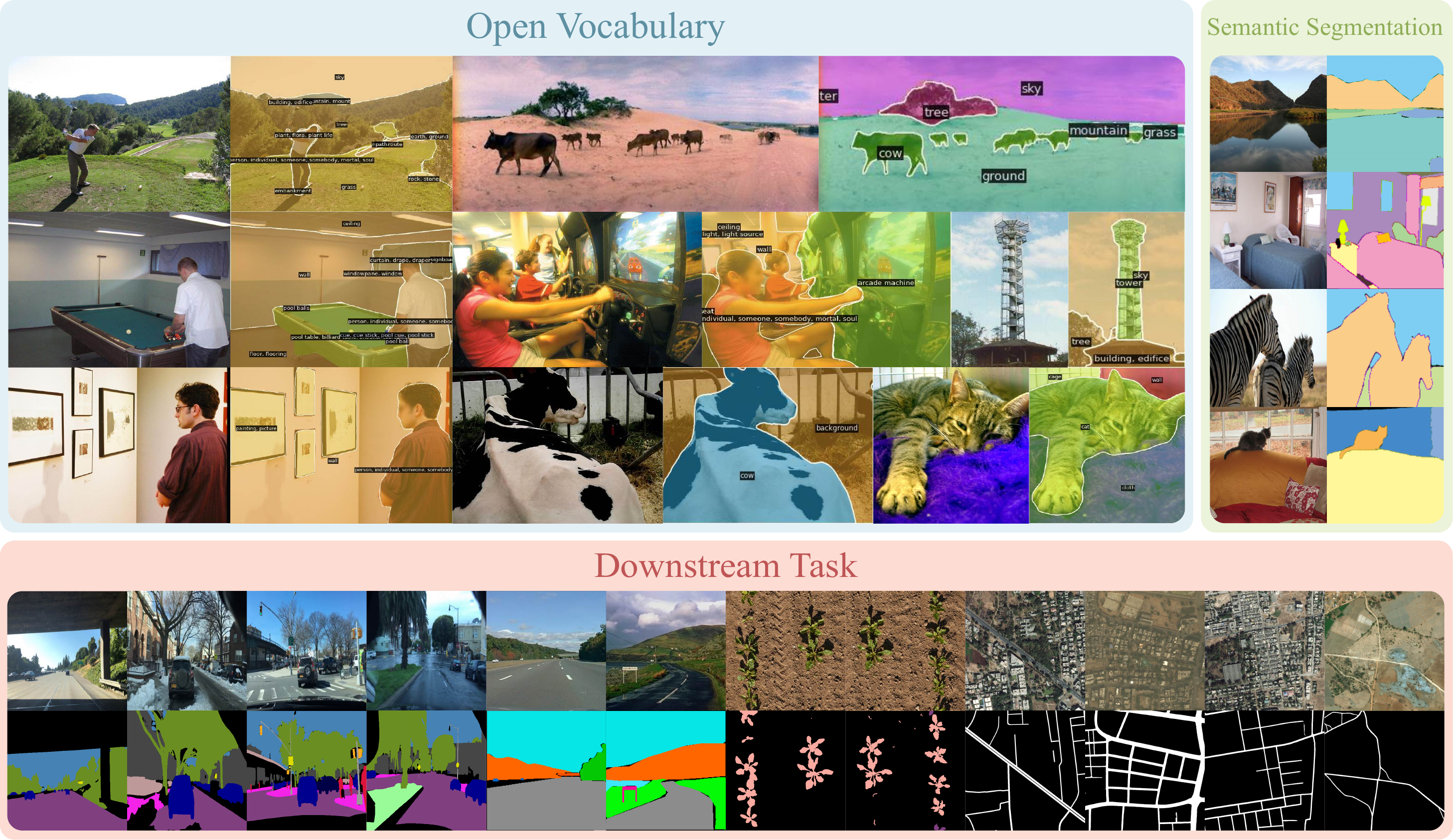} 
    \captionof{figure}{We introduce DiGSeg, a general-purpose segmentation framework built on pretrained diffusion models. By exploiting the strong spatial priors encoded in generative models and fine-tuning with segmentation-aware objectives, DiGSeg produces consistent, high-quality masks across diverse settings, including semantic segmentation, open-vocabulary queries, and cross-domain datasets spanning medical, agricultural, and remote sensing imagery.} 
    \label{fig:teaser}
}

\makeatletter
\apptocmd{\@maketitle}{\centering\insertteaser}{}{}
\makeatother

\begin{document}

\title{Diffusion Model as a Generalist Segmentation Learner} 

\titlerunning{Diffusion Model as a Generalist Segmentation Learner}

\author{
Haoxiao Wang$^{*,\dagger,1}$ \and
Antao Xiang$^{*,2}$ \and
Haiyang Sun$^{*,1}$ \and
Peilin Sun$^{3}$ \and
\\ Changhao Pan$^{1}$ \and
Yifu Chen$^{1}$ \and
Minjie Hong$^{1}$ \and
Weijie Wang$^{1}$ \and
Shuang Chen$^{1}$ \and
\\ Yue Chen$^{4}$ \and
Zhou Zhao$^{\ddagger,1}$
}

\authorrunning{H.~Wang et al.}
\institute{
$^1$ Zhejiang University \quad
$^2$ South China University of Technology \\
$^3$ Nanjing University \quad
$^4$ Peking University \\
Project Page: \url{https://wang-haoxiao.github.io/DiGSeg/} \\
\email{\{haoxiao.wang, zhaozhou\}@zju.edu.cn}
}

\maketitle
\renewcommand{\thefootnote}{\fnsymbol{footnote}}
\footnotetext[1]{Equal contribution, \textsuperscript{$\dagger$} Project Leader, \textsuperscript{$\ddagger$} Corresponding author.}
\renewcommand*{\thefootnote}{\arabic{footnote}}

\begin{abstract}
Diffusion models are primarily trained for image synthesis, yet their denoising trajectories encode rich, spatially aligned visual priors.
In this paper, we demonstrate that these priors can be utilized for text-conditioned semantic and open-vocabulary segmentation, and this approach can be generalized to various downstream tasks to make a general-purpose diffusion segmentation framework.
Concretely, we introduce DiGSeg (\textbf{Di}ffusion Models as a \textbf{G}eneralist \textbf{Seg}mentation Learner), which repurposes a pretrained diffusion model into a unified segmentation framework. Our approach encodes the input image and ground-truth mask into the latent space and concatenates them as conditioning signals for the diffusion U-Net. A parallel CLIP-aligned text pathway injects language features across multiple scales, enabling the model to align textual queries with evolving visual representations.
This design transforms an off-the-shelf diffusion backbone into a universal interface that produces structured segmentation masks conditioned on both appearance and arbitrary text prompts.
Extensive experiments demonstrate state-of-the-art performance on standard semantic segmentation benchmarks, as well as \textbf{strong open-vocabulary generalization} and \textbf{cross-domain transfer} to medical, remote sensing, and agricultural scenarios—without domain-specific architectural customization.
These results indicate that modern diffusion backbones, can serve as generalist segmentation learners rather than pure generators, narrowing the gap between visual generation and visual understanding.
\end{abstract}  

\section{Introduction}
\label{sec:intro}



Modern segmentation systems have achieved remarkable progress across a wide range of tasks—from semantic and instance segmentation in natural scenes to highly specialized domains such as medical imaging, remote sensing, and agriculture \cite{zhou2024image, xu2024advances, huang2023deep}. Despite this progress, the ecosystem remains fundamentally fragmented. Most models are tailored to a specific task or domain, relying on different architectures, label spaces, and training pipelines \cite{zhao2017pyramid, su2025motion, su2025dspv2, li2023mask, he2017mask, wang2023cut}. As a result, transitioning from closed-vocabulary semantic segmentation to open-vocabulary recognition, or from natural images to aerial or medical imagery, often requires redesigning large portions of the system \cite{wang2024improved, yao2024cnn, li2024review}.
This fragmentation highlights a core question: can we design a single, unified segmentation model capable of operating robustly across tasks, vocabularies, and visual domains?

Recent research indicates that diffusion models may provide a pathway to a unified segmentation framework. Several approaches utilize pretrained text-to-image diffusion models by extracting their cross- or self-attention maps to create segmentation masks \cite{van2024simple, zhao2025diception, huang2025vid2world, chen2026learning, le2024maskdiff, zhu2024unleashing, wang2025diffusion, baranchuk2021label, amit2021segdiff}. These studies highlight an intriguing characteristic: diffusion backbones encode rich semantic correspondences that naturally align visual regions with textual or structural cues. 
DiffSeg \cite{tian2024diffuse} aggregates self-attention maps; DiffCut \cite{couairon2024diffcut} segments images by employing diffusion features in conjunction with graph-based clustering; and DiffuMask \cite{wu2023diffumask} leverages cross-attention to synthesize images and masks simultaneously.
Collectively, these methods illustrate that diffusion models organize visual concepts across spatial dimensions, making them a promising foundation for more comprehensive segmentation learning systems.

Despite the potential of diffusion-based repurposing methods, they often fail to deliver reliable, high-quality segmentation results. Many of these methods depend on raw attention maps, which tend to be noisy, low-resolution, and inconsistent across different layers. As a result, they frequently produce fragmented masks that require significant post-processing to be usable \cite{tian2024diffuse, cai2025freemask}. While high-resolution attention maps can provide more detail, they often lack coherence. In contrast, low-resolution maps offer semantic consistency, but at the expense of losing information about small objects and fine boundaries. 
Further analysis of diffusion transformers reveals that although some layers exhibit strong semantic grounding, effectively utilizing them for dense prediction remains a challenge \cite{kim2025seg4diff}. Moreover, existing methods generally focus on a single task—such as open-vocabulary segmentation, panoptic parsing, or instance segmentation—and do not demonstrate that a single diffusion backbone can effectively generalize across multiple tasks and visual domains \cite{xu2023open, gu2024diffusioninst, kim2025distilling, le2024maskdiff}.
In other words, the field currently lacks a unified and conditioned interface that can convert a pretrained diffusion model into a comprehensive segmentation engine.
 
In this paper, we make that step. 
Our main idea is to transform this implicit capability into an explicit segmentation interface using a straightforward fine-tuning protocol. We encode an RGB image along with its ground-truth segmentation map into the latent space of a pretrained diffusion model. In this process, we keep almost all components frozen and fine-tune only the denoising U-Net to generate denoising outputs that align with the segmentation-consistent latents.
This approach retains the powerful visual prior of the generator sourced from a vast internet-scale dataset. Unlike attention-based methods, which rely on post-processing to create masks, our method directly teaches the model to produce high-quality, text-controllable masks from supervised data. Consequently, we develop a diffusion model that functions effectively as a segmentation model rather than just a generator that sometimes produces usable masks.

Our main contributions are as follows:
\begin{itemize}
    \item We present a novel and effective method for fine-tuning a pretrained diffusion model into a segmentation model, showcasing the potential of diffusion models for segmentation tasks.
    \item We introduce a visual latent pathway plus a CLIP-aligned text conditioner that injects language at multiple denoising scales, giving the diffusion U-Net an explicit mechanism to bind textual queries to evolving mask latents.
    \item Our extensive experiments show state-of-the-art performance on standard segmentation tasks, strong zero-shot and open-vocabulary capabilities, and excellent cross-domain transfer without task-specific architectures, proving the model as a generalist segmentation learner.
\end{itemize}

\section{Related Work}
\label{sec:related_workf}

\subsection{Standard Segmentation Tasks}
Image segmentation is a fundamental task in computer vision aimed at dividing an image into pixel-level segments based on semantics \cite{chen2017rethinking, he2017mask, kirillov2019panoptic, strudel2021segmenter, cheng2022masked}.
This involves providing a segmentation mask and a class label for each segment. 
In semantic segmentation, the goal is to output a single segment for each class present in the image \cite{shan2024open, jain2023semask}.
Traditional architectures such as DeepLabv3+ \cite{chen2018encoder} demonstrate the strong ability of CNNs to capture multi-scale context and achieve high accuracy in semantic segmentation benchmarks.
Recently, transformer-based frameworks have further advanced the field. 
Mask2Former \cite{cheng2022masked} adopts a masked-attention transformer decoder to address panoptic, instance, and semantic segmentation in a unified architecture. 
OneFormer \cite{jain2023oneformer} further extends this idea by introducing task-conditioned joint training, enabling a single model to handle all segmentation tasks simultaneously.
Despite these advances, both CNN- and transformer-based models still suffer from limited generalization. They are restricted to finite, dataset-specific vocabularies—far smaller than the rich concepts used to describe the real world—and thus struggle to recognize or segment unseen categories without retraining.

\subsection{Open-Vocabulary Segmentation}
Open-vocabulary segmentation aims to recognize and segment arbitrary categories beyond the training set by leveraging language supervision \cite{du2022learning, ghiasi2022scaling, gu2021open, li2022grounded, minderer2022simple, zareian2021open, li2022language, xu2022groupvit, ghiasi2022scaling}.
Early methods \cite{bucher2019zero, xian2019semantic, zhao2017open}in this area focus on aligning visual features with pre-trained text embeddings by learning a feature mapping that associates visual and text spaces effectively.
With the advent of vision–language models such as CLIP \cite{radford2021learning}, open-vocabulary segmentation has emerged as a promising direction to overcome the limited label spaces of traditional models.
CLIP-based methods like OpenSeg \cite{ghiasi2022scaling}, ZegFormer \cite{ding2022decoupling}, and ZSseg \cite{xu2022simple} typically adopt a two-stage design: class-agnostic masks are first proposed and then matched to CLIP text embeddings for classification.
While ODISE \cite{xu2023open} leverages diffusion models to produce high-quality masks, it still relies on region proposal networks trained with limited annotations and prompt-based category matching. 
In contrast, our model integrates segmentation awareness and text conditioning directly within the diffusion process, removing the need for proposals or handcrafted prompts and achieving stronger generalization across unseen categories and domains.

\subsection{Diffusion Models for Segmentation}
Recent research has explored the potential of using generative models, particularly diffusion models \cite{ho2020denoising}, for segmentation tasks \cite{van2024simple, zhao2025diception, le2024maskdiff, zhu2024unleashing, wang2025diffusion, baranchuk2021label, chen2026unify, amit2021segdiff}. Early efforts, such as DiffuseSeg \cite{tian2024diffuse}, DiffCut \cite{wang2023cut}, DiffuMask \cite{wu2023diffumask}, and Seg4Diff \cite{kim2025seg4diff}, repurposed pretrained text-to-image diffusion models by extracting their internal attention maps or latent features. These approaches demonstrated that diffusion backbones inherently encode strong spatial and semantic organization. 
However, they generally rely on attention maps from intermediate layers, which are often low-resolution, noisy, and inconsistent, resulting in fragmented masks. This leads to a heavy dependency on heuristic post-processing techniques.
In contrast, we propose a more direct and unified approach. Rather than interpreting diffusion attention post-hoc, we fine-tune the denoising U-Net to explicitly produce segmentation-consistent latents.
This design effectively teaches diffusion to segment, generating high-quality, text-controllable masks without the need for manual post-processing.
Our framework recasts the diffusion model from serving solely as a generative prior to functioning as a generalized segmentation learner, preserving its rich visual knowledge while enabling explicit semantic control.


\subsection{Diffusion Models for Dense Prediction}
Akin to segmentation, diffusion models have recently shown great promise in other dense prediction tasks, most notably in monocular depth estimation \cite{ke2024repurposing, duan2024diffusiondepth, patni2024ecodepth, chen2024eqvafford, saxena2023monocular, feng2025seeing, tosi2024diffusion, saxena2023surprising, wang2025transdiff}. 
DiffusionDepth \cite{duan2024diffusiondepth} firstly transforms the monocular depth estimation task into an iterative denoising diffusion process guided by image features.
Marigold \cite{ke2024repurposing} repurposes pretrained image diffusion models to predict continuous depth maps by fine-tuning the denoising U-Net on large-scale depth datasets. 
While both DiGSeg and these depth-oriented models leverage the strong spatial priors of generative backbones, they differ in several key aspects.
First, depth estimation typically focuses on predicting a single-channel continuous value from visual cues alone, whereas DiGSeg is designed as a general-purpose model that handles diverse discrete label spaces—ranging from closed-vocabulary semantic classes to arbitrary open-vocabulary text queries.
Second, unlike those depth estimation models which are primarily image-conditioned, DiGSeg introduces a CLIP-aligned text pathway that enables explicit language-visual grounding at multiple denoising scales. This allows our model to not only capture geometric structures but also to align semantic concepts across diverse domains such as medical and remote sensing imagery, a capability not addressed in specialized depth diffusion models.

\section{Method}
\newcommand{\img}{\bm{x}}
\newcommand{\segment}{\bm{y}}
\newcommand{\latent}{\bm{z}}
\newcommand{\noise}{\bm{\bm{\epsilon}}}
\begin{figure*}[t]
\centering
\includegraphics[width=1.0\textwidth]{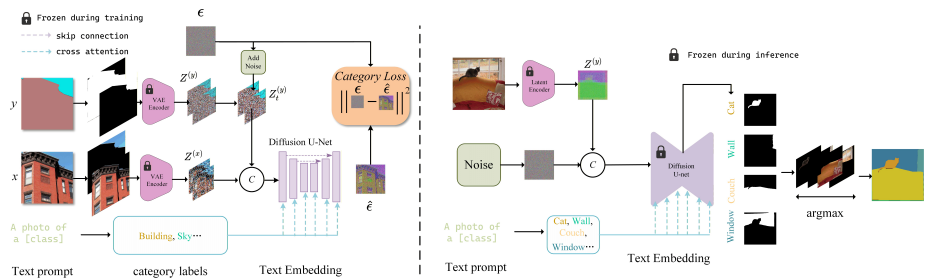} 
\caption{\textbf{DiGSeg pipeline overview}, which presents the training and inference pipelines of our diffusion-based generation model.
In training, paired images are encoded into latent space and, together with text prompts, guide the diffusion U-Net to predict noise using an MSE objective.
In inference, noise is sampled and progressively denoised under text conditioning, after which the VAE decoder reconstructs the final image.}
\label{fig2}
\end{figure*}

\subsection{Problem Definition}
Segmentation tasks can be categorized into three main types: semantic segmentation, open-vocabulary segmentation, and domain-specific segmentation.
Semantic segmentation involves a fixed set of labels, denoted as \(\mathcal{C}_{\text{train}}\), where the goal is to predict a class ID for each pixel in an image. 
Open-vocabulary segmentation expands upon \(\mathcal{C}_{\text{train}}\) to include unseen categories, represented as \(\mathcal{C}_{\text{test}} \neq \mathcal{C}_{\text{train}}\), by utilizing image-text alignment techniques \cite{tu2023open, xu2023open, ghiasi2022scaling}.
Domain-specific segmentation, which can include applications such as medical imaging or remote sensing, often depends on specialized architectures. This reliance can restrict the ability to generalize across different domains.

\subsection{Method Overview}
Figure~\ref{fig2} presents an overview of our DiGSeg framework.
Inspired by previous works \cite{xu2023open, ke2024repurposing}, We found that fine-tuning the diffusion model yields better results than treating the diffusion model as a feature extractor for the segmentation task.
We repurpose a pretrained diffusion model into a unified segmentation learner capable of semantic and open-vocabulary segmentation across diverse domains.
The system contains three key components.
First, the \textbf{Visual Latent Pathway} (Sec.~\ref{sec:3.4}) encodes the RGB image and its segmentation map into a compact latent space using the Stable Diffusion VAE.
This preserves spatial structure while enabling efficient learning of fine-grained correspondences.
Second, the \textbf{CLIP-Aligned Text Conditioning Module} (Sec.~\ref{sec:3.5}) injects language features across denoising steps via a frozen CLIP text encoder, enabling open-vocabulary reasoning without task-specific prompts or additional heads.
Finally, the \textbf{Segmentation-Consistent Denoising U-Net} (Sec.~\ref{sec:3.6}) is trained to denoise toward segmentation-consistent latents conditioned on both image and text features, allowing the diffusion backbone to directly generate segmentation maps instead of relying on attention-based heuristics.
Once trained, DiGSeg supports open-vocabulary inference by conditioning on arbitrary text inputs and generalizes robustly to unseen categories and domains.
The following sections detail each component.

\subsection{Generative Formulation}
We formulate segmentation as a conditional denoising diffusion generation task.
Given an RGB image $\img \in \mathbb{R}^{H \times W \times 3}$ and its corresponding segmentation map $\segment \in \mathbb{R}^{H \times W}$, our goal is to model the conditional distribution $p_\theta(\segment|\img)$ within the latent space of a pretrained diffusion model.

Following the standard diffusion framework, we define a \textit{forward} noising process that gradually corrupts the ground-truth segmentation latent $\segment_0 := \segment$ with Gaussian noise:

\begin{equation}
y_t = \sqrt{\bar{\alpha}_t}\, y_0 + \sqrt{1 - \bar{\alpha}_t}\, \bm{\epsilon}, 
\quad \bm{\epsilon} \sim \mathcal{N}(0, I),
\end{equation}

where $\bar{\alpha}_t = \prod_{s=1}^t (1 - \beta_s)$ and $\{\beta_t\}_{t=1}^T$ denotes the variance schedule over $T$ diffusion steps.
The reverse process is parameterized by a denoising network $\bm{\epsilon}_\theta(\cdot)$, a U-Net, which learns to progressively remove noise from $y_t$ conditioned on $x$:

\begin{equation}
p_\theta(y_{t-1} | y_t, x) = 
\mathcal{N}\!\big(y_{t-1}; \mu_\theta(y_t, x, t), \Sigma_\theta(y_t, x, t)\big).
\end{equation}

During training, parameters $\theta$ are optimized using the standard diffusion objective:

\begin{equation}
\mathcal{L}_{\text{diff}} =
\mathbb{E}_{y_0, \bm{\epsilon}, t}
\big[\|\bm{\epsilon} - \bm{\epsilon}_\theta(y_t, x, t)\|_2^2 \big],
\end{equation}

Where $\hat{\noise} = \noise_{\theta}(\segment_t, \img, t) $ is the noise estimate.
Unlike prior works that model $p_\theta(x|y)$ for image generation, our formulation reverses the conditioning to directly generate segmentation-consistent latents $y$ from the image $\img$. 
At inference time, the segmentation latent $\segment := \segment_0$ is reconstructed starting from a normally distributed variable $\segment_T$, by iteratively applying the learned denoiser $\bm{\epsilon}_\theta(\segment_t, \img, t)$. 
Conditioned on the input image $\img$, the model gradually refines $\segment_t$ through the reverse diffusion process to produce a segmentation-consistent latent representation, which can be decoded into the final mask prediction $\segment_0$.

\subsection{Visual Latent Pathway}
\label{sec:3.4}
The encoder $\mathcal{E}$ compresses both the RGB image and its corresponding segmentation map into compact latent representations:
\[
z_x = \mathcal{E}(x), \quad z_y = \mathcal{E}(y),
\]
where $z_x, z_y \in \mathbb{R}^{h \times w \times c}$ denote the encoded latent features of the image and the segmentation map, respectively. The decoder $\mathcal{D}$ allows reconstruction back to the pixel space, i.e., $\hat{x} = \mathcal{D}(z_x)$ and $\hat{y} = \mathcal{D}(z_y)$, ensuring perceptual consistency between latent and data domains.Because the pretrained VAE is designed for 3-channel RGB inputs, the single-channel segmentation map is replicated across three channels to simulate an RGB image before encoding.
This simple strategy maintains compatibility with the encoder without retraining or modifying the latent structure.


\subsection{Text Conditioner}
\label{sec:3.5}
To endow our model with open-vocabulary and text-controllable segmentation capability, we introduce a CLIP-Aligned Text Conditioner that injects language information into the diffusion process. At each denoising timestep, this module provides semantic grounding by aligning textual and visual representations within the U-Net.

Given a class name or a natural-language description, we obtain a text embedding $t_{\text{clip}}$ using a frozen CLIP text encoder. This embedding is then integrated into multiple denoising scales of the U-Net through cross-attention:
\begin{equation}
h_t = \text{UNet}(z_t, t_{\text{clip}}, t),
\end{equation}
where $z_t$ denotes the noisy latent at timestep $t$. Multi-scale language injection ensures that both global semantics and local spatial cues are jointly refined during denoising.



\subsection{Segmentation-Consistent Denoising}
\label{sec:3.6}
Given the encoded latent pairs $(z_x, z_y)$ from the Visual Latent Pathway, we randomly sample a timestep $t \in [1, T]$ and add Gaussian noise to the segmentation latent $z_y$ according to the forward diffusion process:
\begin{equation}
z_y^t = \sqrt{\bar{\alpha}_t} \, z_y + \sqrt{1 - \bar{\alpha}_t}\, \epsilon, \quad \epsilon \sim \mathcal{N}(0, I)
\end{equation}

where $\bar{\alpha}_t$ follows the predefined noise schedule. The denoiser, implemented as a U-Net $\epsilon_\theta$, then learns to predict and remove this noise conditioned on both the image latent $z_x$ and the text embedding $t_{\text{clip}}$:
\begin{equation}
\mathcal{L}_{\text{seg}} = 
\mathbb{E}_{t, \epsilon}\big[\|\epsilon - \epsilon_\theta(z_y^t, z_x, t, t_{\text{clip}})\|_2^2\big].
\end{equation}

\paragraph{Noise Strategy.}
To stabilize training and accelerate convergence, we employ an \textit{annealed multi-scale noise schedule} \cite{song2020improved, Whitaker2023MultiResolution}. 
Instead of adding noise purely at a single resolution, we combine Gaussian noise components at multiple spatial scales, each upsampled to match the latent resolution.
At early diffusion steps, high-frequency perturbations dominate to encourage fine-structure learning; as denoising progresses, low-frequency components become more influential, gradually emphasizing semantic structure.
This annealed multi-scale perturbation improves spatial coherence and yields smoother, more accurate segmentation boundaries.

\subsection{Inference}
At inference time, our model performs conditional latent diffusion denoising to generate segmentation maps from an input image.  
Given an RGB image $x$, we first encode it into the latent space as $z_x = \mathcal{E}(x)$.  
A segmentation latent $z_y^T$ is then initialized as standard Gaussian noise $z_y^T \sim \mathcal{N}(0, I)$ and progressively denoised under the same schedule used during training:

\begin{equation}
z_y^{t-1} = \frac{1}{\sqrt{\alpha_t}}\Big(z_y^t - \frac{1-\alpha_t}{\sqrt{1-\bar{\alpha}_t}} \, \epsilon_\theta(z_y^t, z_x, t, t_{\text{clip}})\Big) + \sigma_t \epsilon,
\end{equation}

where $\sigma_t$ controls the noise level and $\epsilon_\theta$ denotes our segmentation-consistent denoiser.  
After completing all reverse steps, the clean segmentation latent $z_y^0$ is decoded back to the pixel space using the frozen VAE decoder:
\begin{equation}
\hat{y} = \mathcal{D}(z_y^0).
\end{equation}


\paragraph{Test-Time Ensembling.}
Given the stochastic nature of the diffusion process, we optionally adopt a lightweight test-time ensemble: multiple inference passes are run with different noise seeds, and the resulting segmentation maps are averaged in the latent space before decoding.  
This aggregation improves spatial consistency and reduces noise artifacts, particularly in fine-structure regions, with minimal additional computational cost.

\begin{table*}[!h]
\begin{center}
\resizebox{\textwidth}{!}{
\begin{tabular}{l|cc|cc|ccccc}
\toprule
\textbf{Method} & \textbf{VLM} & \textbf{Backbone} & \textbf{Training Dataset} & \textbf{Additional Dataset} & \textbf{A-847} & \textbf{PC-459} & \textbf{A-150} & \textbf{PC-59} & \textbf{Cityscapes} \\
\midrule\midrule
ODISE \cite{xu2023open} & CLIP ViT-L/14 & Stable Diffusion & COCO-Panoptic & \xmark & 11.0 & 13.8 & 28.7 & 55.3 & - \\
OVSeg \cite{liang2023open} & CLIP ViT-L/14 & Swin-B  & COCO-Stuff & \checkmark & 9.0 & 12.4 & 29.6 & 55.7 & - \\
SAN \cite{xu2023side} & CLIP ViT-L/14 & Side Adapter & COCO-Stuff & \xmark & 12.4 & 15.7 & 29.9 & 51.8 & - \\
SCAN \cite{liu2024open} & CLIP ViT-L/14 & Swin-B & COCO-Stuff & \xmark & 14.0 & 16.7 & 33.5 & 59.3 & - \\
CAT-Seg \cite{cho2024cat} & CLIP ViT-L/14 & - & COCO-Stuff & \xmark & 16.0 & 23.8 & 31.5 & 62.0 & - \\
MAFT+ \cite{jiao2024collaborative} & ConvNeXt-L & - & COCO-Stuff & \xmark & 15.1 & 21.6 & 36.1 & 59.4 & - \\
SED \cite{xie2024sed} & CLIP ConvNeXt-L & - & COCO-Stuff & \xmark & 13.9 & 22.6 & 35.2 & 60.6 & - \\
Mask-Adapter \cite{li2025mask} & CLIP ConvNeXt-L & - & COCO-Stuff & \xmark & 16.2 & 22.7 & 38.2 & 60.4 & \underline{37.9} \\
Seg4Diff \cite{kim2025seg4diff} & CLIP ViT-L/14 & Stable Diffusion & COCO-Stuff & \xmark & - & - & 35.2 & 51.2 & 26.0 \\
HyperCLIP \cite{peng2025understanding} & CLIP ViT-L/14 & - & COCO-Stuff & \xmark & 16.3 & 24.1 & 38.2 & 64.2 & - \\
OVSNet \cite{liu2025stepping} & CLIP ViT-L/14 & ResNet-101c & COCO-Stuff & \xmark & 16.2 & 23.5 & 37.1 & 62.0 & - \\
DPSeg \cite{zhao2025dpseg} & CLIP ConvNeXt-L & - & COCO-Stuff & \xmark & 15.7 & 24.1 & 37.1 & 62.3 & - \\
SemLA \cite{qorbani2025semantic} & CLIP ConvNeXt-L & - & COCO-Stuff & \xmark & - & - & 36.9 & 62.2 & - \\
ESC-Net \cite{lee2025effective} & CLIP ViT-L/14 & - & COCO-Stuff & \xmark & \underline{18.1} & \underline{27.0} & \underline{41.8} & \underline{65.6} & - \\
\hlrow
& & & & & \textbf{19.9} & \textbf{29.2} & \textbf{43.2} & \textbf{68.4} & \textbf{38.5} \\
\hlrow \multirow{-2}{*}{\textbf{DiGSeg (ours)}} 
 & \multirow{-2}{*}{CLIP ViT-L/14}
 & \multirow{-2}{*}{Stable Diffusion}
 & \multirow{-2}{*}{COCO-Stuff}
 & \multirow{-2}{*}{\xmark}
 & \textcolor{ForestGreen}{(+1.8)}
 & \textcolor{ForestGreen}{(+2.2)}
 & \textcolor{ForestGreen}{(+1.4)}
 & \textcolor{ForestGreen}{(+2.8)}
 & \textcolor{ForestGreen}{(+0.6)} \\
\midrule[1pt]

OVSeg \cite{liang2023open} & CLIP ViT-B/16 & ResNet-101c & COCO-Stuff & \checkmark & 7.1 & 11.0 & 24.8 & 53.3 & - \\
SCAN \cite{liu2024open} & CLIP ViT-B/16 & Swin-B & COCO-Stuff & \xmark & 10.8 & 13.2 & 30.8 & 58.4 & - \\
EBSeg \cite{shan2024open} & CLIP ViT-B/16 & SAM ViT-B & COCO-Stuff & \xmark & 11.1 & 17.3 & 30.0 & 56.7 & - \\
SED \cite{xie2024sed} & ConvNeXt-B & - & COCO-Stuff & \xmark & 11.4 & 18.6 & 31.6 & 57.3 & - \\
CAT-Seg \cite{cho2024cat} & CLIP ViT-B/16 & - & COCO-Stuff & \xmark & 12.0 & 19.0 & 31.8 & 57.5 & - \\
OPMapper \cite{wangopmapper} & CLIP ViT-B/16 & Swin-B & COCO-Stuff & \xmark & - & - & 31.0 & 58.3 & - \\
ESC-Net \cite{lee2025effective} & CLIP ViT-B/16 & - & COCO-Stuff & \xmark & 13.3 & \underline{21.1} & \underline{35.6} & \underline{59.0} & - \\
HyperCLIP \cite{peng2025understanding} & CLIP ViT-B/16 & - & COCO-Stuff & \xmark & 12.3 & 19.2 & 32.1 & 58.5 & - \\
Mask-Adapter \cite{li2025mask} & CLIP ConvNeXt-B & - & COCO-Stuff & \xmark & \underline{14.2} & 17.9 & \underline{35.6} & 58.4 & \underline{35.2} \\
DPSeg \cite{zhao2025dpseg} & CLIP ConvNeXt-B & - & COCO-Stuff & \xmark & 12.5 & 20.1 & 33.3 & 58.4 & - \\
\hlrow
& & & & & \textbf{17.5} & \textbf{23.1} & \textbf{37.2} & \textbf{62.7} & \textbf{36.5} \\
\hlrow \multirow{-2}{*}{\textbf{DiGSeg (ours)}} 
 & \multirow{-2}{*}{CLIP ViT-B/16}
 & \multirow{-2}{*}{Stable Diffusion}
 & \multirow{-2}{*}{COCO-Stuff}
 & \multirow{-2}{*}{\xmark}
 & \textcolor{ForestGreen}{(+3.3)}
 & \textcolor{ForestGreen}{(+2.0)}
 & \textcolor{ForestGreen}{(+1.6)}
 & \textcolor{ForestGreen}{(+3.7)}
 & \textcolor{ForestGreen}{(+1.3)} \\



\bottomrule
\end{tabular}
}
\caption{\textbf{Quantitative evaluation on open-vocabulary segmentation benchmarks} All methods are evaluated on A-847, PC-459, A-150, PC-59, and Cityscapes. We highlight the \textbf{best}, while the second-best results are underlined. Improvements over the second-best are highlighted in \textcolor{ForestGreen}{green}.}
\label{tab:open-vocabulary}
\end{center}
\end{table*}

\paragraph{Hyperparameter $\tau$ Tuning.}
Since the output of our diffusion model is a continuous-valued mask in $[0,1]$, a threshold $\tau$ is required to convert the predicted logits into a binary mask for each class. 
The choice of $\tau$ can influence the sharpness and coverage of the resulting mask, and different categories may exhibit different optimal threshold values depending on object size, texture, and spatial sparsity. 
\begin{wrapfigure}[9]{r}{0.5\textwidth}
    \centering
    \vspace{-20pt}
    \includegraphics[width=1.0\linewidth]{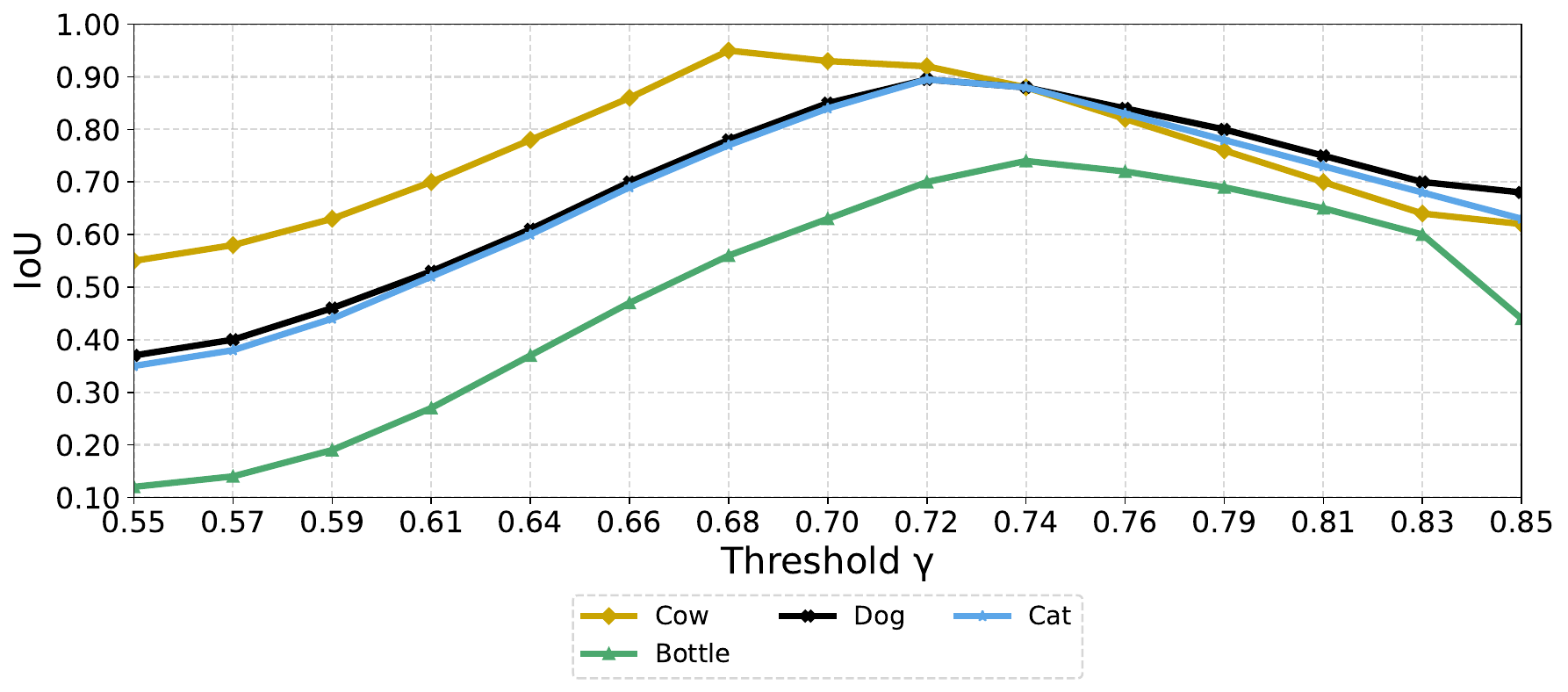}
    \vspace{-0.22in}
    \caption{\textbf{Impact of Hyperparameter $\tau$.}}
    \label{supply-1}
\end{wrapfigure}
To characterize this behavior, we examine how IoU varies with respect to $\tau$ for several representative categories on Fig.~\ref{supply-1}.
The optimal threshold can shift across classes, with small or fine-grained objects typically favoring slightly lower thresholds, while larger and more homogeneous objects prefer higher thresholds.
We intentionally avoid post-processing to preserve the generality and simplicity of our model. 
Our empirical findings indicate that a single fixed value of $\tau = 0.7$ consistently delivers strong performance across semantic segmentation, open-vocabulary segmentation, and all downstream segmentation benchmarks.

\begin{table}[t]
\centering
\renewcommand{\arraystretch}{1.1} 
\begin{minipage}{0.53\textwidth}
 \centering
 \setlength{\tabcolsep}{1pt} 
 \resizebox{\linewidth}{!}{ 
 \begin{tabular}{lcccc}
\toprule
\multirow{2}{*}{\textbf{Method}} & \multicolumn{2}{c}{\textbf{COCO}} & \multicolumn{2}{c}{\textbf{ADE20K}} \\
\cmidrule(lr){2-3} \cmidrule(lr){4-5}
                        & input size     & miou    & input size      & miou     \\
\midrule\midrule
SegFormer-B5 \cite{xie2021segformer}      & $512^2$ & 46.7 & $640^2$ & 51.8 \\
Mask2Former-Swin-L \cite{cheng2022masked} & - & -  & $640^2$  & \underline{57.3} \\
OneFormer \cite{jain2023oneformer}         & -  & -  & $640^2$  & 57.0 \\
LDMSeg  \cite{van2024simple}           & - & -  & $512^2$  & 52.2 \\
PEM \cite{cavagnero2024pem}               & -  & -  &  $512^2$  & 45.5 \\
FeedFormer-B2 \cite{shim2023feedformer}        & -  & -  &  $512^2$  & 48.0 \\
CGRSeg-L \cite{ni2024context}          &  $512^2$ & 46.0 &  $512^2$ & 48.3 \\
VWFormer-B5 \cite{yan2024multi}       &  $512^2$ & 48.0  &  $512^2$ & 54.7 \\
SegMAN \cite{fu2025segman}            &  $512^2$ & 48.2  &  $512^2$ & 53.2 \\
EoMT \cite{kerssies2025your}              & $512^2$  & \underline{48.7} & $512^2$  & 57.1 \\
OffSeg-L \cite{zhang2025revisiting}          & $512^2$  & 46.0  & $512^2$ & 48.5 \\
MambaVision-B \cite{hatamizadeh2025mambavision}     & -  & -  & $512^2$  & 49.1 \\
\hlrow
& & \textbf{50.8} & & \textbf{58.6} \\
\hlrow \multirow{-2}{*}{\textbf{DiGSeg (ours)}} 
  & \multirow{-2}{*}{$512^2$}
  & \textcolor{ForestGreen}{(+2.1)}
  & \multirow{-2}{*}{$512^2$}
  & \textcolor{ForestGreen}{(+1.3)} \\
\bottomrule
\end{tabular}
 }
 \caption{\textbf{Quantitative evaluation on semantic segmentation.} Our DiGSeg demonstrates outstanding performance.}
 \label{tab:segmentation_comparison}
\end{minipage}
\hfill 
\begin{minipage}{0.45\textwidth}
\centering
\resizebox{\linewidth}{!}{
\begin{tabular}{l|cccc}
\toprule
\textbf{Method} & \textbf{IoU$_{road}$} & \textbf{Precision} & \textbf{Recall} & \textbf{F1} \\
\midrule\midrule
\textcolor{gray}{DDCTNet} \cite{gao2024ddctnet}     
& \textcolor{gray}{64.27} & \textcolor{gray}{79.02} & \textcolor{gray}{78.10} & \textcolor{gray}{78.24} \\
\textcolor{gray}{Unet}  \cite{ronneberger2015u}       
& \textcolor{gray}{62.94} & \textcolor{gray}{-} & \textcolor{gray}{-} & \textcolor{gray}{-} \\
\textcolor{gray}{D-LinkNet} \cite{zhou2018d}   
& \textcolor{gray}{63.00} & \textcolor{gray}{-} & \textcolor{gray}{-} & \textcolor{gray}{-} \\
\textcolor{gray}{CoANet} \cite{mei2021coanet}      
& \textcolor{gray}{60.65} & \textcolor{gray}{76.98} & \textcolor{gray}{72.34} & \textcolor{gray}{74.61} \\
\textcolor{gray}{SGCN}  \cite{zhou2021split}       
& \textcolor{gray}{53.92} & \textcolor{gray}{72.95} & \textcolor{gray}{68.25} & \textcolor{gray}{72.31} \\
\textcolor{gray}{DeepLabv3} \cite{wang2024improved}   
& \textcolor{gray}{61.97} & \textcolor{gray}{77.26} & \textcolor{gray}{74.09} & \textcolor{gray}{75.64} \\
\textcolor{gray}{Segroad} \cite{tao2023seg} 
& \textcolor{gray}{66.23} & \textcolor{gray}{-} & \textcolor{gray}{-} & \textcolor{gray}{-} \\
\textcolor{gray}{CGC-Net} \cite{zheng2023cgc} 
& \textcolor{gray}{68.80} & \textcolor{gray}{82.67} & \textcolor{gray}{80.39} & \textcolor{gray}{81.51} \\

\midrule
SegMAN & 48.12 & 70.25 & 68.10 & 69.16 \\
EoMT & 52.52 & 72.88 & 71.35 & 72.10 \\
OffSeg & 51.75 & 72.10 & 70.85 & 71.46 \\
MambaVision & \underline{57.28} & \underline{75.32} & \underline{74.50} & \underline{74.91} \\
\hlrow
& \textbf{65.78} & \textbf{79.93} & \textbf{78.92} & \textbf{78.79} \\
\hlrow \multirow{-2}{*}{\textbf{DiGSeg (ours)}} 
  & \textcolor{ForestGreen}{(+8.50)} 
  & \textcolor{ForestGreen}{(+4.61)} 
  & \textcolor{ForestGreen}{(+4.42)} 
  & \textcolor{ForestGreen}{(+3.88)} \\
\bottomrule
\end{tabular}
}
\caption{\textbf{Quantitative evaluation on DeepGlobe road segmentation tasks.} 
The module specifically designed for remote sensing tasks is denoted by \textcolor{gray}{gray}.
}
\label{tab:road_segmentation_comparison}
\end{minipage}
\end{table}
\section{Experiments}
\label{experiments}
We begin this section by describing the implementation details of our framework.
We then compare our method with SOTA approaches on semantic and open-vocabulary segmentation benchmarks.
Finally, we conduct comprehensive ablation studies to validate the contribution of each component in our model.
Due to page constraints, more implementation details, dataset tests, detailed experimental results and metrics, as well as ablation experiment analyses, are presented in the \textbf{supplementary materials}.

\subsection{Implementation Details}
\paragraph{Architecture}
We implement DiGSeg utilize SDv2 \cite{rombach2022high}.
We adopt the DDPM noise schedule \cite{ho2020denoising} with $T=1000$ steps during training. For each training iteration, we sample a timestep $t \sim \mathcal{U}(\{1,\dots,T\})$.
At inference time, we use a DDIM sampler \cite{song2020denoising} with between 1 and 50 steps.
For all segmentation tasks, unless otherwise stated, we ensemble 8 predictions from different initial noise.
we resize the input images to $512 \times 512$, use random horizontal flip and scale jittering in $[0.8, 1.2]$. 
The model is trained for 30000 iterations with a total batch size of 16.
We use AdamW with a learning rate of $1\times10^{-4}$.
Training and testing are conducted using 8 NVIDIA A100 GPU, For most semantic segmentation and open vocabulary segmentation tasks, it takes approximately 1 days to converge during training.

\begin{figure*}[!t]
    \centering
    \includegraphics[width=1.0\linewidth]{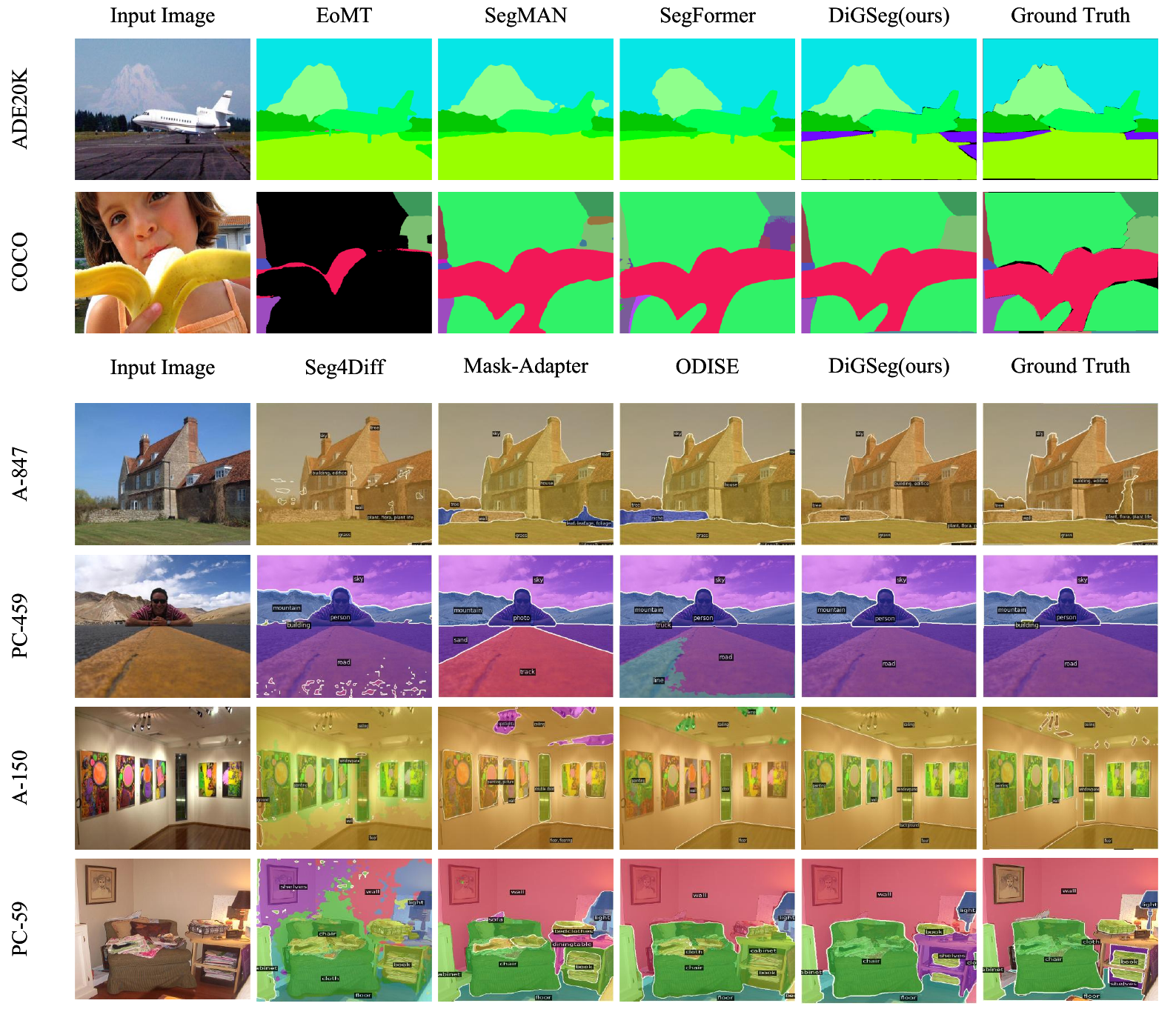}
    \caption{\textbf{Qualitative comparison} of semantic and open-vocabulary segmentation across different datasets.}
    \label{visual-1}
\end{figure*}
\begin{figure*}[!t]
    \centering
    \includegraphics[width=0.95\linewidth]{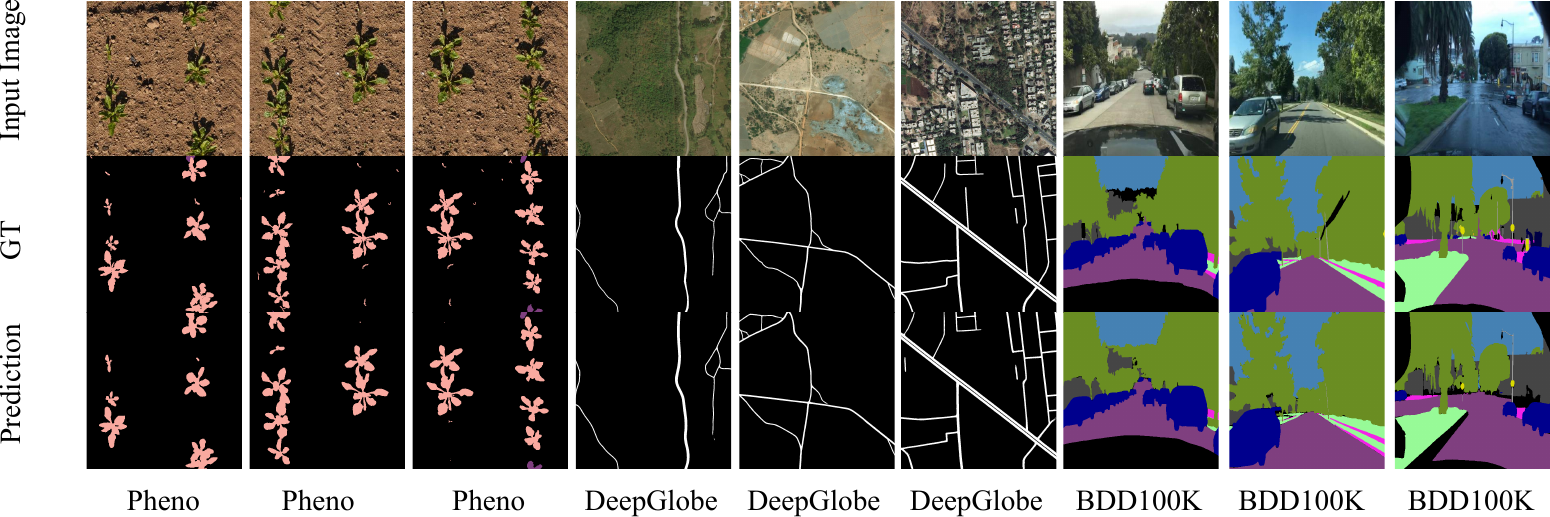}
    \caption{\textbf{Qualitative results} of cross-domain segmentation across different datasets.}
    \label{visual-1}
\end{figure*}

\paragraph{Training datasets}
We use widely adopted benchmarks for image segmentation.
Specifically, we evaluate the semantic segmentation performance on COCO \cite{lin2014microsoft} and ADE20K \cite{zhou2019semantic}.
Following the experiment in \cite{ghiasi2022scaling, xu2023side}, we report performance on A-150 with 150 common classes and A-847 with all the 847 classes of ADE20K \cite{zhou2019semantic}, PC-59 with 59 common classes and PC-459 with full 459 classes of Pascal Context \cite{mottaghi2014role}, we also incorporated the CityScapes \cite{cordts2016cityscapes} dataset in order to conduct a more comprehensive comparison.
For task-specific segmentation tasks, we conducted extensive experiments using various datasets from medical, remote sensing, and agricultural domains. These datasets include Agriculture-Vision \cite{chiu2020agriculture}, REFUGE-2 \cite{fang2022refuge2}, BraTS-2021 \cite{baid2021rsna}, DeepGlobe \cite{demir2018deepglobe}, LoveDA \cite{wang2021loveda}, and BDD100K \cite{yu2020bdd100k}. Due to page constraints, we have focused on presenting the Phenobench \cite{weyler2024phenobench} and DeepGlobe \cite{demir2018deepglobe} datasets in the main sections.

\paragraph{Evaluation metrics}
For semantic and open-vocabulary segmentation, we evaluate performance using the mean Intersection-over-Union (mIoU) across all classes, following the official dataset splits.
For all task-specific segmentation benchmarks, mIoU is used as the primary metric. In the medical segmentation setting, we additionally report the Dice score to better capture region overlap quality.

\subsection{Comparison with State of the Art}
\paragraph{Semantic and open-vocabulary segmentation} We compare DiGSeg with state-of-the-art segmentation frameworks in several segmentation tasks.
As shown in Tab.~\ref{tab:open-vocabulary}, for open-vocabulary segmentation, we compare DiGSeg to thirteen baselines on mIoU score. 
For almost the datasets, we achieved state-of-the-art results.
Notebly, We adopt the public results of integrating Mask-Adapter \cite{li2025mask} into MAFTP \cite{jiao2024collaborative} as the baseline (i.e., 'MAFTP w/ MaskAdapter'), similarly, SemLA \cite{qorbani2025semantic} is based on CAT-Seg \cite{cho2024cat}, and OPMapper \cite{wangopmapper} is based on SCAN \cite{liu2024open}.
The results of semantic segmentation are presented in Tab.~\ref{tab:segmentation_comparison}.
For all baseline methods, we select the variant with the largest number of parameters to ensure a fair comparison.
Our proposed DiGSeg consistently outperforms existing segmentation frameworks on both the COCO and ADE20K benchmarks.
Specifically, DiGSeg achieves an mIoU of \textbf{50.8} on COCO and \textbf{58.6} on ADE20K, exceeding the second-best results by \textbf{+2.1} and \textbf{+1.3}, respectively.

\paragraph{Task-specific segmentation}
The quantitative results for the DeepGlobe dataset can be found in Table~\ref{tab:road_segmentation_comparison}, while the results for the Monuseg dataset are shown in supplementary materials.
To demonstrate the generalized capabilities of our model, all settings are consistent with those used for semantic segmentation.
Notably, our model demonstrates a significant performance in the general domains as well as in agriculture and  remote sensing whitout any change in architecture.
Additionally, the scores in medical domains are relatively low, likely reflecting CLIP’s limited understanding of this specialized field.

\subsection{Ablation Study}

\paragraph{\textbf{Data efficiency under limited supervision.}}

\begin{figure}[!t]
    \centering
    \begin{minipage}{0.48\textwidth}
        \centering
        \includegraphics[width=1.0\linewidth]{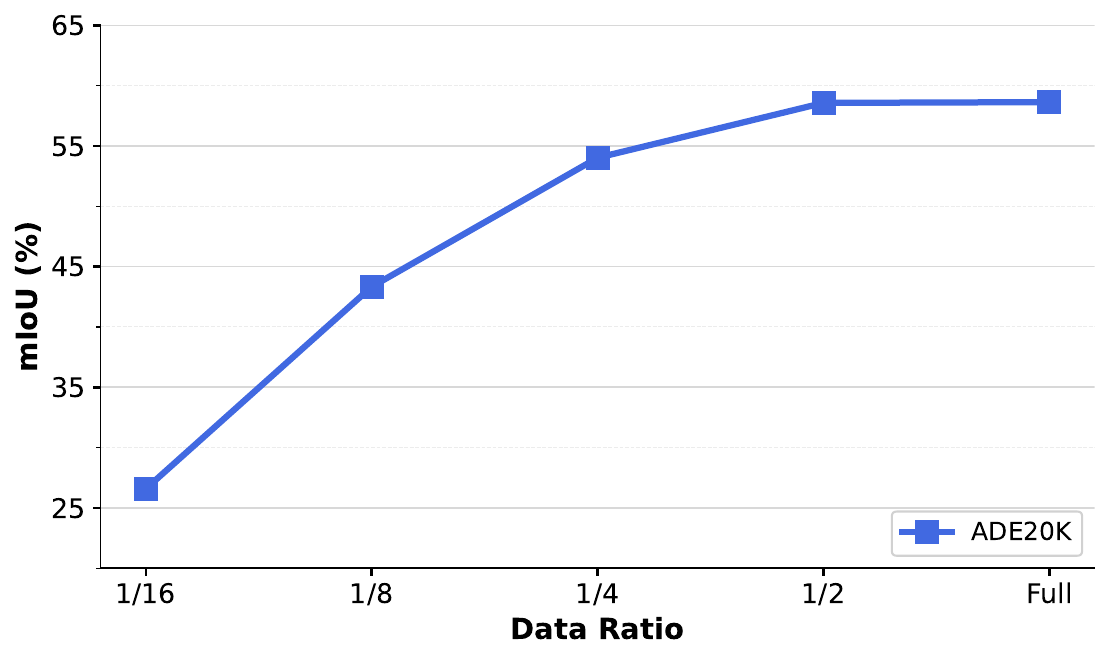}
        \caption{\textbf{Effect of training data ratio.}
        Our model exhibits remarkable data efficiency, maintaining nearly equivalent performance even when trained on only 50\% of the total dataset.}
        \label{fig:ablation-1}
    \end{minipage}
    \hfill 
    \begin{minipage}{0.48\textwidth}
        \centering
        \includegraphics[width=1.0\linewidth]{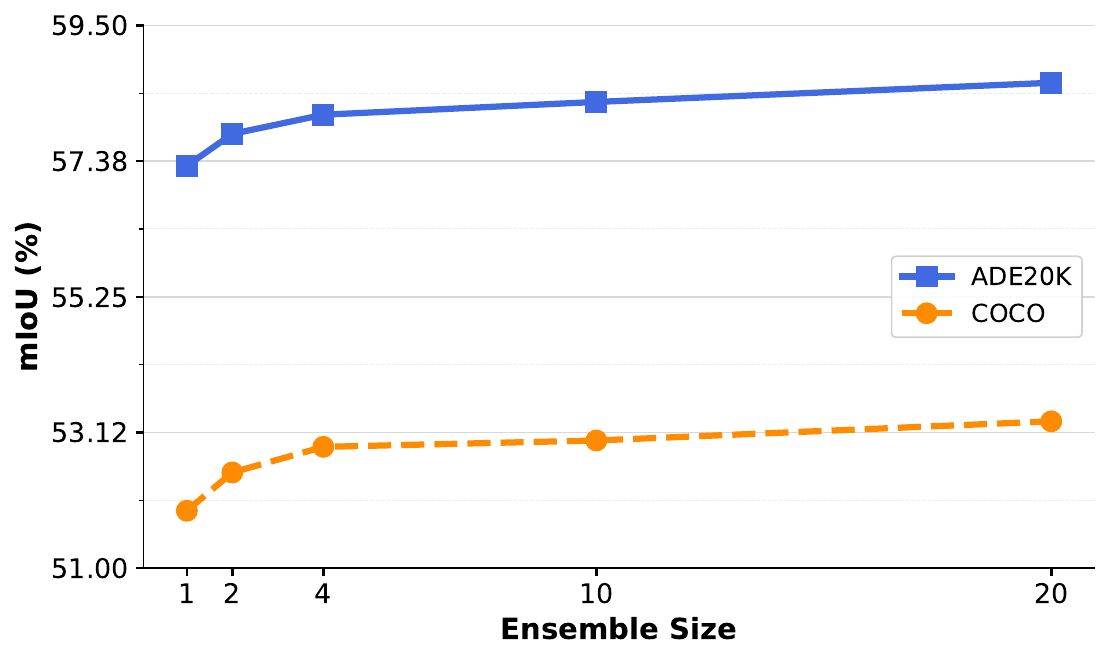}
        \caption{\textbf{Ablation of ensemble size.} We observe a consistent improvement as the ensemble size increases. However, this improvement begins to taper off after 10 predictions per sample.}
        
        \label{fig:ablation-2}
    \end{minipage}
\end{figure}

To examine data efficiency and robustness to limited supervision, we train our model with $1/2$, $1/4$, $1/8$ and $1/16$ of the ADE20K training data.
Figure~\ref{fig:ablation-1} shows that our model can achieve nearly identical performance using only half the amount of data compared to the full dataset.
Furthermore, when trained with just a quarter of the data, the results are still impressively strong.
This shows that our diffusion-based segmentation learner effectively utilizes pretrained visual priors and acquires generalizable representations in low-data scenarios.

\paragraph{\textbf{Test-time ensembling.}}
We investigated the impact of the number of test-time ensembling iterations on the outcome.
As shown in Figure~\ref{fig:ablation-2}, a single prediction has already been able to yield reasonably good results. Integration ten times will result in a 
1.2\% improvement on ADE20K dataset and 1.9\% improvement on COCO dataset.

\begin{figure}[!t]
    \centering
    \includegraphics[width=0.9\linewidth]{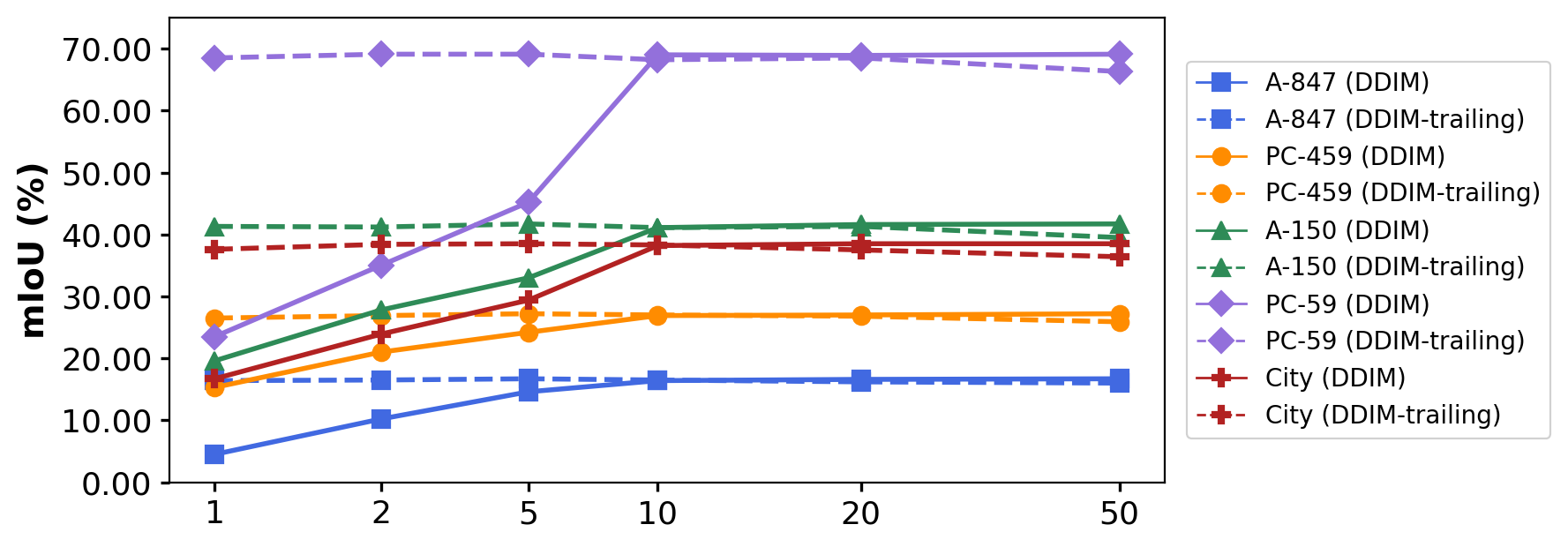}
    \caption{\textbf{Ablation study of denoising schedule.}
    Under the DDIM-trailing setting, one step of denoising can achieve a relatively high effect, while DDIM requires at least 10 steps.}
    \label{fig:ablation-3}
\end{figure}

\vspace{-10pt}
\paragraph{\textbf{Denoising steps.}}
In dense prediction scenarios, \cite{garcia2025fine} observed that the conventional denoising diffusion implicit models (DDIM) leading time-step strategy causes inconsistencies between the training and inference processes.
Based on this finding, we will switch to using trailing timesteps for the DDIM.
This change has significantly improved our model's efficiency, with performance reaching saturation after just 1 DDIM step.
Figure~\ref{fig:ablation-3} shows the effect of applying DDIM-trailing on the open-vocabulary benchmark.

\vspace{-10pt}
\paragraph{\textbf{Training noise.}}
We compared the effects of four different types of noise on the training process. 
As shown in Table~\ref{tab:ablation_noise_schedule}, we discovered that incorporating multi-resolution noise alone led to a significantly greater improvement compared to using standard Gaussian noise.
In contrast, adding annealing noise by itself had a relatively minor effect when applied to standard Gaussian noise.
The best results were achieved by applying annealing noise after introducing multi-resolution noise.

\begin{table}[!t]
\centering
\renewcommand{\arraystretch}{1.1}

\begin{minipage}[t]{0.52\linewidth}
 \centering
 \resizebox{\linewidth}{!}{
 \begin{tabular}{l|cc}
 \toprule
 \textbf{Method} & \textbf{COCO} & \textbf{ADE20K} \\
 \midrule
 Standard Gaussian noise & 48.9 & 56.7 \\
 \midrule
 w/ Annealed noise & 49.2 & 57.1 \\
 w/ Multi-resolution noise & 49.7 & 57.6 \\
 \midrule
 \rowcolor{gray!10} 
 \textbf{w/ Multi-res. + Ann. (ours)} & \textbf{50.8} & \textbf{58.6} \\
 \bottomrule
 \end{tabular}
 }
 \caption{\textbf{Ablation on noise schedule.} Combining multi-resolution and annealed noise yields best performance.}
 \label{tab:ablation_noise_schedule}
\end{minipage}
\hfill 
\begin{minipage}[t]{0.45\linewidth}
 \centering
 \resizebox{\linewidth}{!}{
 \begin{tabular}{ll|c|cc}
 \toprule
 \textbf{COCO} & \textbf{ADE20K} & \textbf{E-Step} & \textbf{Citys.} & \textbf{BDD100K} \\
 \midrule
 \multirow{2}{*}{\xmark} & \multirow{2}{*}{\checkmark} 
 & $1\times1$ & 41.22 & 37.55 \\
 & & $8\times2$ & \textbf{43.58} & \textbf{37.64} \\
 \midrule
 \multirow{2}{*}{\checkmark} & \multirow{2}{*}{\xmark} 
 & $1\times1$ & 37.80 & 35.76 \\
 & & $8\times2$ & 38.51 & 36.42 \\
 \midrule
 \multirow{2}{*}{\checkmark} & \multirow{2}{*}{\checkmark} 
 & $1\times1$ & 38.74 & 36.89 \\
 & & $8\times2$ & 39.04 & 37.21 \\
 \bottomrule
 \end{tabular}
 }
 \caption{\textbf{Cross-domain evaluation.} Results on Cityscapes and BDD100K under different training sets.}
 \label{tab:cross_domain_exs}
\end{minipage}

\end{table}
\vspace{-10pt}



\paragraph{\textbf{Cross-domain analysis.}}

We compare models trained on COCO, ADE20K, and both datasets by evaluating them on Cityscapes and BDD100K. As shown in Table~\ref{tab:cross_domain_exs}, ADE20K training yields the highest mIoU on both targets. We conduct a systematic analysis of the number of ensembles (E) and DDIM denoising steps (S), denoted jointly as the E-Step setting $(E \times S)$. 
This suggests that ADE20K offers greater scene diversity and more detailed pixel-level annotations, which effectively capture the spatial context and semantics of urban environments.
These results confirm that more data or mixed datasets do not necessarily improve cross-domain performance—dataset relevance is crucial.

\paragraph{\textbf{Inference Speed.}}
In Table \ref{tab:ablation_estep}, we present the inference runtime of DiGSeg. As a diffusion-based method, we admit our model is slower than feed-forward segmenters, which reflects the typical trade-off between quality and efficiency. 
\begin{wraptable}[9]{r}{0.5\textwidth}
\vspace{-20pt}
\centering
\renewcommand{\arraystretch}{1.1}
\resizebox{0.90\linewidth}{!}{
\begin{tabular}{l|cc|c}
\toprule
\textbf{E-Step} & \textbf{COCO} & \textbf{ADE20K} & \textbf{FPS} \\
\midrule
1x1                 & 48.2 & 56.8 & 11.27 \\
1x2                 & 48.5 & 57.1 & 10.61 \\
4x2                 & 50.5 & 58.5 & 5.82  \\
8x2                 & 50.8 & 58.6 & 3.15  \\
20x50 (w/o trailing) & 50.9 & 58.8 & 0.12 \\
\bottomrule
\end{tabular}
}
\vspace{-5pt}
\caption{\textbf{Ablation on E-Step scheduling.}
Increasing the E-Step range improves segmentation accuracy but reduces inference speed.}
\label{tab:ablation_estep}
\end{wraptable}
However, many established diffusion acceleration techniques can be applied to further enhance the speed of future versions of our framework. In this work, we implement DDIM-trailing, which significantly improves upon standard diffusion sampling, making DiGSeg noticeably faster than previous diffusion-based segmentation models.

\section{Conclusion}
\vspace{-10pt}
In this paper, we presented \textbf{DiGSeg}, a novel framework that successfully repurposes pretrained diffusion models from pure generators into generalized segmentation learners. By introducing a latent-conditioned fine-tuning protocol and a multi-scale CLIP-aligned text pathway. Our extensive evaluations across standard benchmarks and specialized domains demonstrate that DiGSeg not only achieves state-of-the-art performance but also exhibits exceptional open-vocabulary generalization. These results confirm that the rich visual priors within diffusion backbones can be effectively channeled through a unified interface, bridging the critical gap between generative modeling and dense visual understanding.

\clearpage
\appendix
\noindent
In this appendix, we provide additional implementation details in~\cref{sec:supp_implementation}, present extended ablation studies in~\cref{sec:ablation}, and include further quantitative and qualitative results in~\cref{sec:supp_quan} and~\cref{sec:supp_qualitative}, respectively.

\section{More Implementation Details}
\label{sec:supp_implementation}
\subsection{Multi-Resolution Noise Construction}

Following the multi-scale noising strategy commonly used in latent diffusion models, we construct a noise pyramid 
$\{\epsilon^{(i)}\}_{i=0}^{L}$, 
where each level corresponds to a progressively lower spatial resolution.
Noise at level $i$ is produced by downsampling a base Gaussian tensor, sampling independent Gaussian noise at the low resolution, and then upsampling to match the latent resolution.
The standard multi-resolution formulation combines these signals with exponentially decaying weights $s^{i}$, where $0 < s < 1$ controls the global influence of low-frequency noise components.
To make the multi-resolution noise closer to the Gaussian distribution assumed in original DDPM theory while still benefiting from multi-scale structure, we apply \textit{annealing} to progressively reduce the effect of low-resolution components as the denoising approaches the final steps.
Specifically, for diffusion timestep $t \in \{1, \dots, T\}$, the noise at pyramid level $i > 0$ is assigned a time-dependent weight $w_i(t) = \left( s \frac{t}{T} \right)^{i}$
where $T$ is the total number of diffusion steps.

\subsection{Training Architecture Details}
DiGSeg is built on the SDv2 U-Net backbone, utilizing its pretrained multi-scale convolutional and attention structures while keeping the VAE encoder and decoder frozen to maintain consistency in the latent space. 
We fine-tune only the cross-attention blocks at the 1/4 and 1/8 resolutions, the mid-block attention, and a small MLP projection that is used to inject CLIP ViT-L/14 text embeddings. All other convolutional layers adhere to the original SD configuration and remain frozen. 
Text embeddings are integrated through cross attention at two spatial scales, providing class guidance during each denoising run.

\subsection{Inference Details}
Since the main paper focuses on the diffusion-based learning formulation, we include here the full protocol for (1) constructing candidate class lists, (2) conditioning the model with text embeddings, and (3) aggregating per-class predictions into a final dense segmentation map.
\paragraph{Class label generation.}
We employ a CLIP-based open-vocabulary classifier to propose a compact set of likely classes for each image.
For every benchmark, we define a fixed vocabulary of valid semantic categories, and we obtain a text embedding for each class $c$ using the prompt \texttt{"A photo of a {class}"}.
The CLIP text encoder produces a normalized embedding $t_c$, and all ${t_c}$ are precomputed and cached.
Given an input image $x$, we extract its CLIP image embedding $v = \mathrm{CLIP}_{\text{img}}(x)$ and measure cosine similarity to all class embeddings, yielding scores $s_c = \cos(v, t_c)$.
Classes with similarity above a threshold $\tau$ (default $0.25$) are retained as initial candidates.
To better recover small or localized objects that may not strongly influence the global embedding, we further incorporate a patch-level refinement.
We collect patch tokens ${v_p}$ from the CLIP image encoder and compute their similarity to $t_c$.
If any patch exhibits a strong local response, $\max_p \cos(v_p, t_c) > \tau_{\text{patch}}$, the corresponding class is added to the candidate set regardless of its global score.
This patch-aware refinement reliably retrieves small or low-visibility categories and complements the global filtering step, yielding a more complete and robust per-image class list.

\paragraph{Mask generation}
For each candidate class $c \in \mathcal{C}(x)$, we first retrieve its precomputed text embedding $t_c$, inject $t_c$ into the cross-attention layers of the diffusion U\,-Net, run DDIM-trailing inference to denoise from the VAE latent representation of the image, and finally decode the predicted latent into a continuous mask representation $\hat{M}_c \in \mathbb{R}^{H \times W}$. Thus, each denoising run produces a class-specific mask logit map, formally expressed as $\hat{M}_c = f_\theta(x, t_c)$, where $\hat{M}_c(i,j)$ denotes the confidence that pixel $(i,j)$ belongs to class $c$.
\paragraph{Mask Aggregation}
After predicting masks for all candidate class, we aggregate them into a final dense segmentation map. We first collect all predicted mask logits as $\mathcal{M}(x)=\{\hat{M}_c \mid c \in \mathcal{C}(x)\}$. For each pixel location $(i,j)$, we assign its label by taking a pixel-wise argmax over all candidate classes, i.e., $\hat{y}(i,j)=\arg\max_{c \in \mathcal{C}(x)}\hat{M}_c(i,j)$. If the maximum logit response at a pixel is below a background threshold $\tau_{\mathrm{bg}}$, we instead assign it to a background label, namely $\hat{y}(i,j)=\text{background}$ if $\max_{c}\hat{M}_c(i,j)<\tau_{\mathrm{bg}}$. This simple rule empirically works well for all benchmarks.

\section{More Ablation Experiments}
\label{sec:ablation}
\paragraph{SD VAE with Segmentation Masks}
To evaluate how well the Stable Diffusion VAE can faithfully handle segmentation masks, we conduct a quantitative study on ADE20K and COCO-Stuff. 
For each dataset, we randomly sample 1,000 ground-truth masks, expand the single-channel label maps into three channels, and normalize them to match the VAE’s input range. The masks are passed through the pretrained VAE, and the decoded RGB outputs are converted back to discrete labels via nearest-neighbor palette matching.
We measure reconstruction quality using pixel-wise Mean Absolute Error (MAE). Across all datasets, the error is extremely small, i.e., $0.0041 \pm 0.0037$ on a set of segmentation maps, far below common segmentation noise thresholds and occupying less than $0.5\%$ of the VAE output range $[-1, 1]$. As shown in Tab.~\ref{tab:vae_channel_consistency}, this suggests that the VAE preserves the sharp, piecewise-constant structure of segmentation masks with minimal distortion.
These findings confirm that the Stable Diffusion VAE is expressive enough to encode segmentation masks without degrading boundaries or label semantics, enabling effective learning in the latent space.

\begin{table*}[!t]
\centering
\renewcommand{\arraystretch}{1.1}
\resizebox{0.5\textwidth}{!}{
\begin{tabular}{l|cc}
\toprule
\textbf{Dataset} & \textbf{std} & \textbf{max -- min} \\
\midrule
ADE20K     & 0.0018 & 0.0055 \\
COCO       & 0.0022 & 0.0067 \\
Cityscapes & 0.0027 & 0.0081 \\
VOC        & 0.0024 & 0.0076 \\
\bottomrule
\end{tabular}
}
\caption{\textbf{Channel-wise consistency after VAE decoding.} 
The discrepancy across the three reconstructed RGB channels remains extremely small for all datasets.}
\label{tab:vae_channel_consistency}
\vspace{-6pt}
\end{table*}

\section{More Quantitative results}
\label{sec:supp_quan}
In this section, we presented the additional results on the downstream task datasets.
It is worth noting that the performance in the medical field was not satisfactory, which might be due to the fact that CLIP has insufficient understanding of medical data.
The results of Pheno-Bench Dataset is shown on Tab.~\ref{pheno}(a) and the results of REFUGE-2 Dataset is shown on Tab.~\ref{pheno}(b).
\subsection{Results on Pheno-Bench Dataset}
The Pheno-Bench dataset focuses on fine-grained plant phenotyping, requiring the model to distinguish between subtle morphological variations among plant organs (e.g., leaves, stems, and fruits). As shown in Tab.~\ref{pheno}(a), our method achieves competitive performance compared with existing baselines, demonstrating that the learned representations can transfer effectively to agricultural and biological imaging domains. The improvement is particularly notable in categories with high intra-class variability, suggesting that our framework captures more discriminative semantic cues beyond what vanilla CLIP features provide.

\subsection{Results on REFUGE-2 Dataset}
The REFUGE-2 dataset targets the segmentation of optic disc and optic cup regions in retinal fundus images, which is critical for glaucoma diagnosis. As reported in Tab.~\ref{pheno}(b), although our method outperforms several strong baselines, the absolute performance in this medical domain is comparatively limited. We attribute this gap primarily to the inherent domain mismatch: CLIP is pre-trained on web-scale natural image-text pairs and therefore exhibits insufficient understanding of medical imagery, where visual cues differ significantly from natural scenes. Furthermore, medical terminology in the text encoder space is sparsely represented, leading to weaker text-image alignment for clinical concepts.

\begin{table}[!t]
\centering
\caption{\textbf{Quantitative evaluations on Phenobench and REFUGE-2 datasets.}}
\label{tab:combined_results}
\begin{minipage}{0.54\linewidth}
\centering
\footnotesize
\renewcommand{\arraystretch}{0.92}
\resizebox{\linewidth}{!}{
\begin{tabular}{l|ccc}
\toprule
\textbf{Method} & \textbf{mIoU} & \textbf{IoU$_{crop}$} & \textbf{IoU$_{weed}$} \\
\midrule
\textcolor{gray}{SSWM} \cite{celikkan2023semantic}
& \textcolor{gray}{78.49} 
& \textcolor{gray}{94.60} 
& \textcolor{gray}{63.37} \\
\midrule
SegMAN      & 71.31 & 84.22 & 58.41 \\
EoMT        & 71.82 & 86.20 & 57.45 \\
OffSeg      & \underline{74.85} & \underline{88.91} & \underline{60.79} \\
MambaVision & 73.91 & 87.78 & 60.05 \\
\hlrow
& \textbf{76.66} & \textbf{90.94} & \textbf{62.38} \\
\hlrow \multirow{-2}{*}{\textbf{DiGSeg (ours)}} 
  & \textcolor{ForestGreen}{(+1.81)} 
  & \textcolor{ForestGreen}{(+2.03)} 
  & \textcolor{ForestGreen}{(+1.59)} \\
\bottomrule
\end{tabular}
}
\subcaption{\textbf{Quantitative evaluation on Phenobench weed–crop segmentation.}
The module specifically designed for agricultural scenarios is shown in \textcolor{gray}{gray}.}
\end{minipage}
\hfill 
\begin{minipage}{0.44\linewidth}
\centering
\footnotesize
\renewcommand{\arraystretch}{0.92}
\resizebox{\linewidth}{!}{
\begin{tabular}{l|cc}
\toprule
\textbf{Method} & \textbf{IoU} & \textbf{Dice} \\
\midrule
\textcolor{gray}{BEAL} \cite{wang2019boundary}         & \textcolor{gray}{74.1} & \textcolor{gray}{83.5} \\
\textcolor{gray}{ResUnet} \cite{yu2019robust}     & \textcolor{gray}{72.3} & \textcolor{gray}{80.1} \\
\textcolor{gray}{MRNet}  \cite{ji2021learning}      & \textcolor{gray}{75.1} & \textcolor{gray}{84.2} \\
\textcolor{gray}{nnUNet} \cite{isensee2021nnu}  
& \textcolor{gray}{75.1} & \textcolor{gray}{84.3} \\
\textcolor{gray}{MedSegDiff} \cite{wu2024medsegdiff}  & \textcolor{gray}{\underline{79.1}} & \textcolor{gray}{\underline{87.5}} \\
\midrule
SegMAN       & 46.7 & 63.7 \\
EoMT         & 49.9 & 66.6 \\
OffSeg       & \textbf{53.5} & \textbf{69.7} \\
MambaVision  & \underline{50.4} & \underline{67.0} \\
\textbf{DiGSeg (ours)} & 34.5 & 51.3 \\
\bottomrule
\end{tabular}
}
\subcaption{\textbf{Quantitative comparison on REFUGE-2 medical image segmentation.} The module specifically designed for medical scenarios is shown in \textcolor{gray}{gray}.}
\end{minipage}
\label{pheno}
\end{table}

\section{More Qualitative results}
\label{sec:supp_qualitative}
In this section, we show additional qualitative results on open-vocabulary and semantic segmentation tasks.

\subsection{Open-Vocabulary Segmentation}
We first present qualitative comparisons on four widely adopted open-vocabulary segmentation benchmarks, which differ substantially in the number of categories and the granularity of annotations.

\textbf{A-847.} As illustrated in Fig.~\ref{visual-1}, A-847 contains the largest vocabulary (847 categories), posing significant challenges for fine-grained recognition. Our method produces accurate segmentation masks even for rare and visually similar categories, while preserving sharp object boundaries. Compared with baseline approaches, our predictions exhibit fewer category confusions, particularly for small or occluded objects.

\textbf{PC-459.} Fig.~\ref{visual-2} shows results on PC-459, where the model must differentiate between 459 categories with diverse appearances. Our method demonstrates strong robustness in cluttered scenes and successfully segments objects with complex shapes. The qualitative comparisons further reveal that our approach better captures contextual relationships between foreground objects and background regions.

\textbf{A-150.} On the A-150 benchmark (Fig.~\ref{visual-3}), which contains 150 commonly seen categories, our method consistently produces clean and coherent segmentation masks. We observe that the predicted masks align well with object contours, indicating that our framework effectively leverages both semantic and spatial cues.

\textbf{PC-59.} As shown in Fig.~\ref{visual-4}, PC-59 features a relatively smaller vocabulary but still includes diverse indoor and outdoor scenes. Our method generates visually pleasing results with smooth boundaries and accurate region assignments, demonstrating its applicability to general-purpose segmentation scenarios.

\subsection{Semantic Segmentation}
We further evaluate our method on two large-scale semantic segmentation benchmarks to verify its effectiveness under the closed-set setting.

\textbf{ADE20K.} Fig.~\ref{visual-5} presents qualitative results on the ADE20K dataset, which covers a wide range of scene types and object categories. Our method handles complex scene compositions effectively, yielding precise segmentation for both stuff (e.g., sky, road, wall) and thing (e.g., person, car, furniture) classes. Notably, our approach maintains consistent performance across varying object scales and lighting conditions.

\textbf{COCO.} Finally, Fig.~\ref{visual-6} illustrates results on the COCO dataset. Our method produces high-quality segmentation maps with sharp boundaries and accurate class assignments, even in scenarios involving heavy occlusion or significant scale variation. The qualitative results corroborate our quantitative findings and further demonstrate the generalization capability of our framework.

\section{Discussion}
Our results situate DiGSeg within a rapidly growing line of work that repurposes generative models for visual understanding. While generative pretraining historically lagged behind discriminative and self-supervised methods~\cite{chen2020generative, bai2024sequential}, the maturation of large-scale text-to-image diffusion models is changing this picture: Marigold~\cite{ke2024repurposing} and Lotus~\cite{he2024lotus} show that diffusion U-Nets encode geometric priors transferable to depth and surface-normal estimation, while ODISE~\cite{xu2023open}, DiffCut~\cite{couairon2024diffcut}, Seg4Diff~\cite{kim2025seg4diff}, and Diception~\cite{zhao2025diception} explore similar priors for perception. DiGSeg advances this trajectory by directly fine-tuning the denoiser to produce segmentation-consistent latents under multi-scale CLIP-aligned text conditioning, turning the diffusion model from a passive feature extractor into a controllable segmentation engine that generalizes across closed-vocabulary, open-vocabulary, and cross-domain settings.

A particularly notable concurrent development is Vision Banana ~\cite{gabeur2026image}, which pushes this hypothesis to its extreme: by lightly instruction-tuning a frontier image generator to emit RGB visualizations of vision-task outputs, a single generalist model surpasses specialists such as SAM 3 on segmentation and Depth Anything 3 on metric depth, while preserving its generation ability.  We view DiGSeg and Vision Banana as complementary points on the same trajectory, differing mainly in operating point but converging on the same core message: the priors learned during large-scale visual generation are a substrate for general visual understanding, not merely a means to synthesis. The convergence of these works suggests that the boundary between generative and discriminative vision is rapidly dissolving.

\begin{figure*}[!t]
    \centering
    \includegraphics[width=0.90\linewidth]{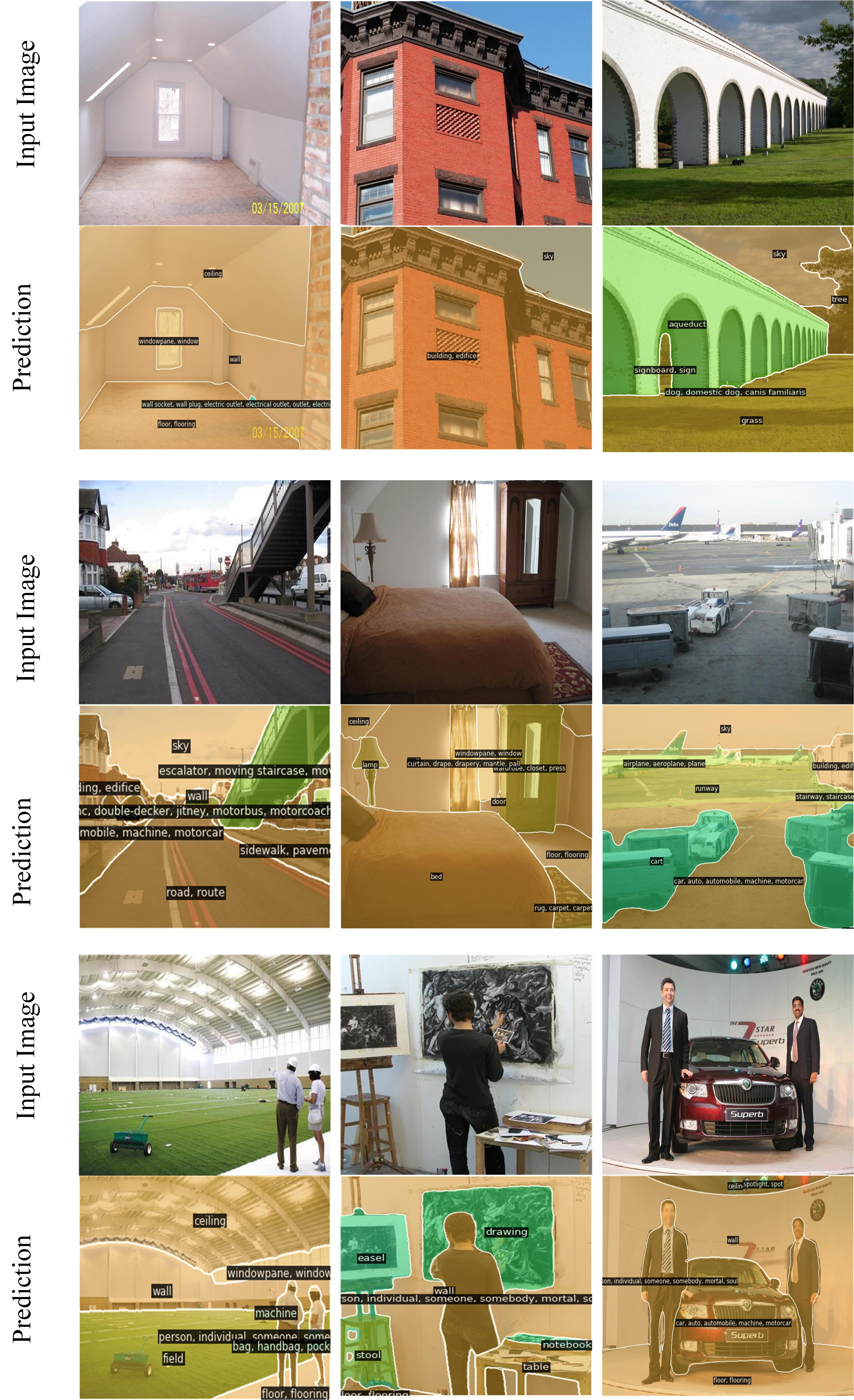}
    \caption{\textbf{Qualitative results} A-847 on open-vocabulary segmentation.}
    \label{visual-1}
\end{figure*}

\begin{figure*}[!t]
    \centering
    \includegraphics[width=0.90\linewidth]{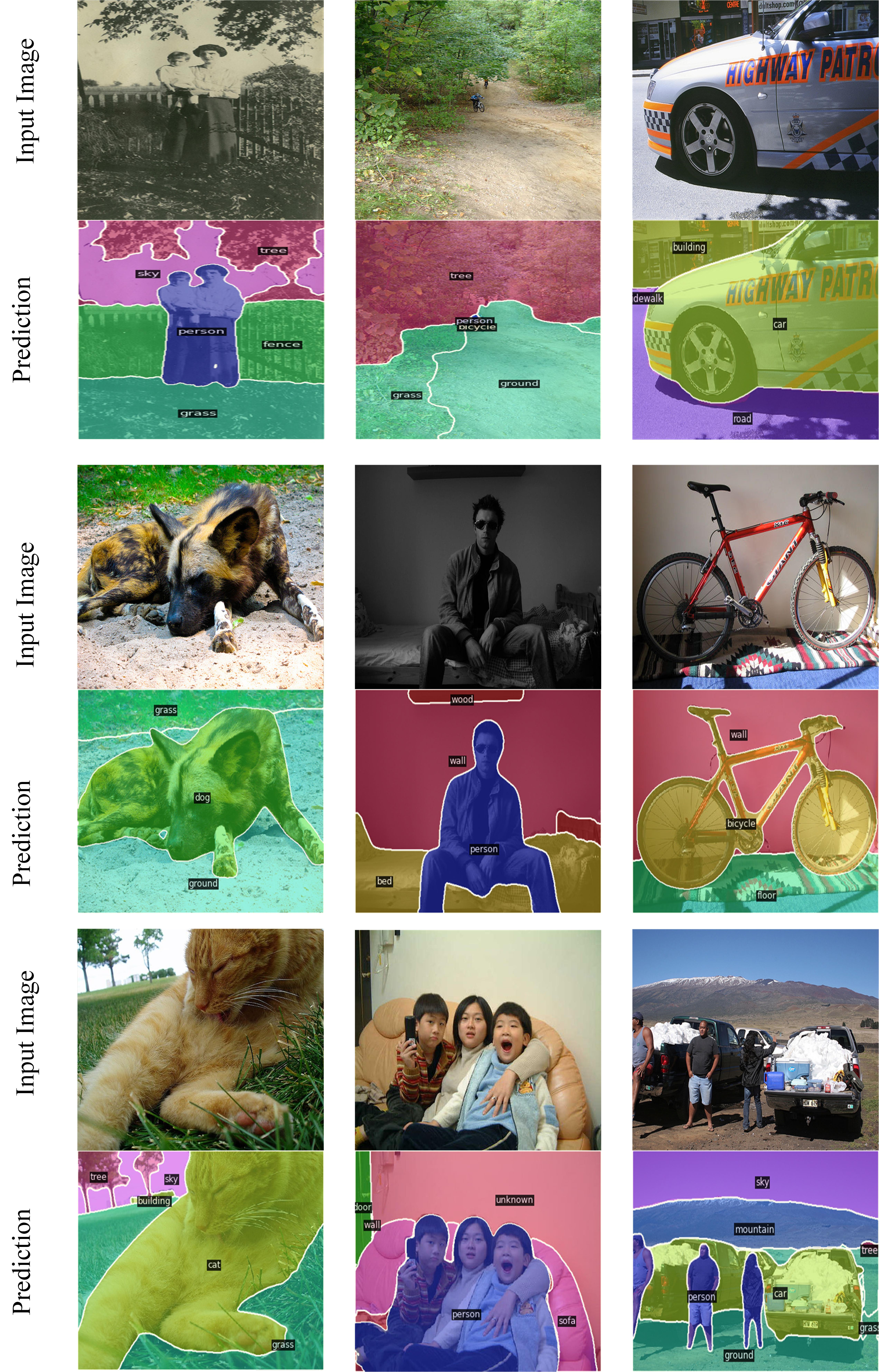}
    \caption{\textbf{Qualitative results} PC-459 on open-vocabulary segmentation.}
    \label{visual-2}
\end{figure*}

\begin{figure*}[!t]
    \centering
    \includegraphics[width=0.90\linewidth]{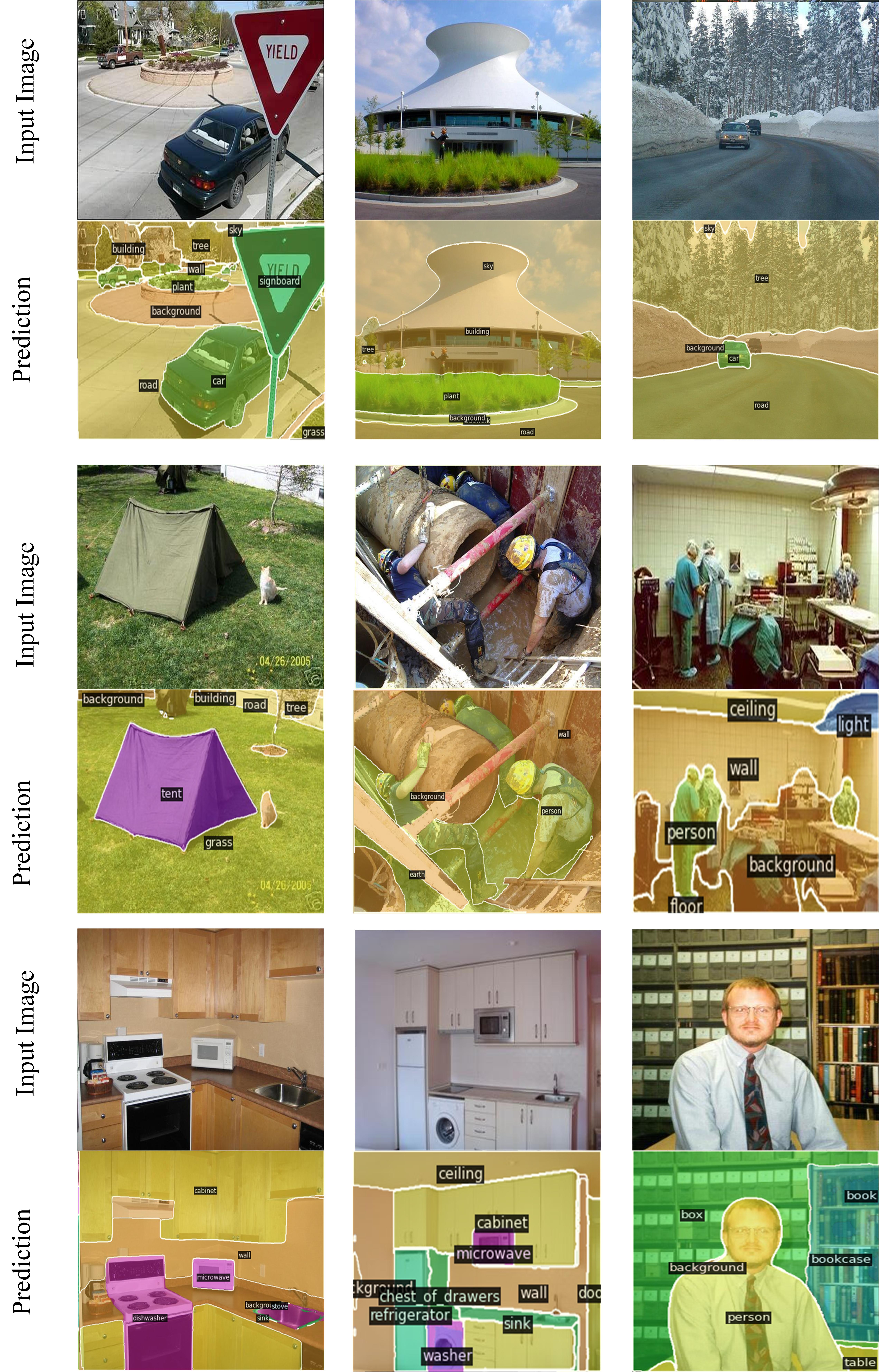}
    \caption{\textbf{Qualitative results} A-150 on open-vocabulary segmentation.}
    \label{visual-3}
\end{figure*}

\begin{figure*}[!t]
    \centering
    \includegraphics[width=0.90\linewidth]{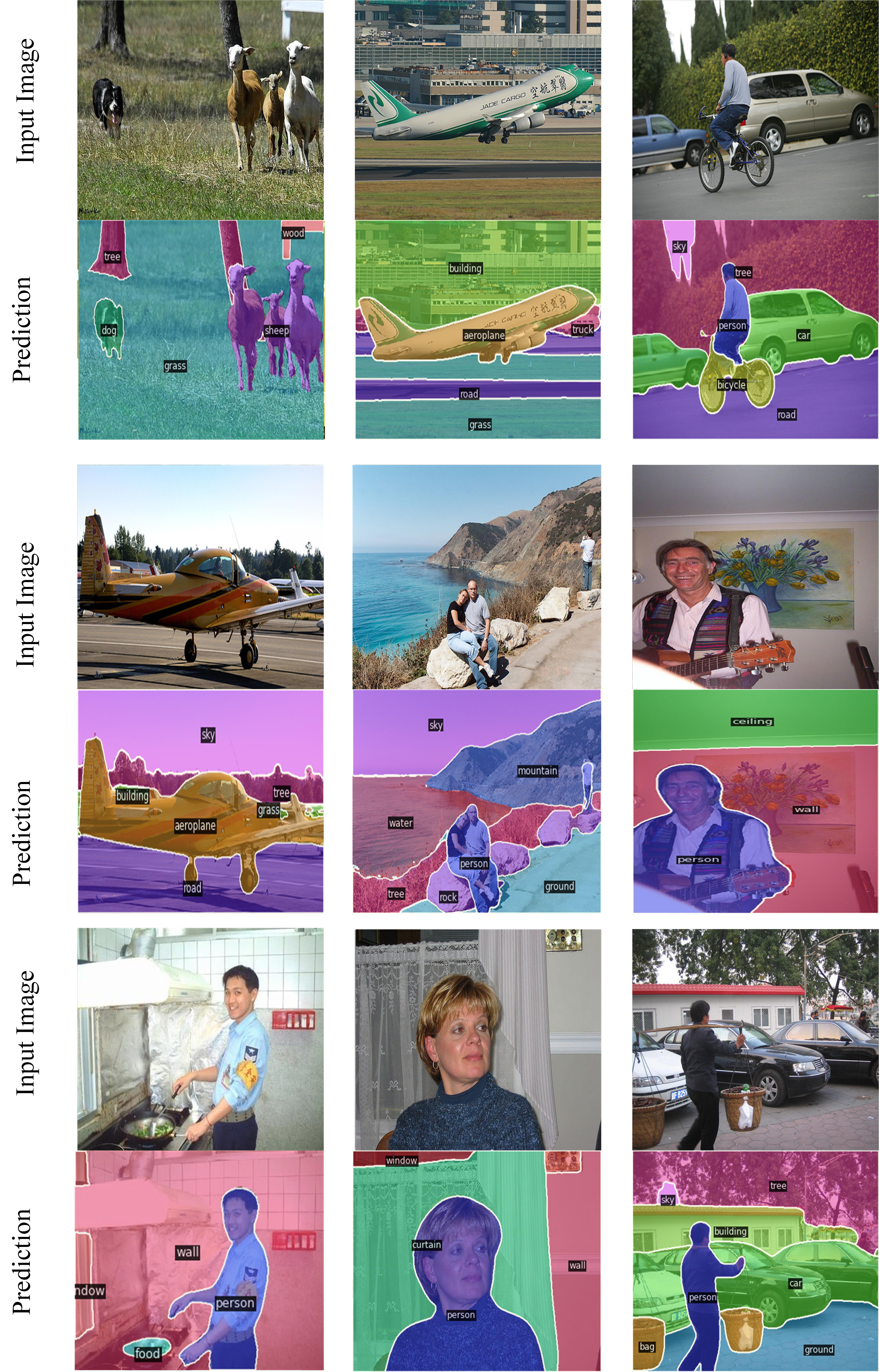}
    \caption{\textbf{Qualitative results} PC-59 on open-vocabulary segmentation.}
    \label{visual-4}
\end{figure*}

\begin{figure*}[!t]
    \centering
    \includegraphics[width=0.90\linewidth]{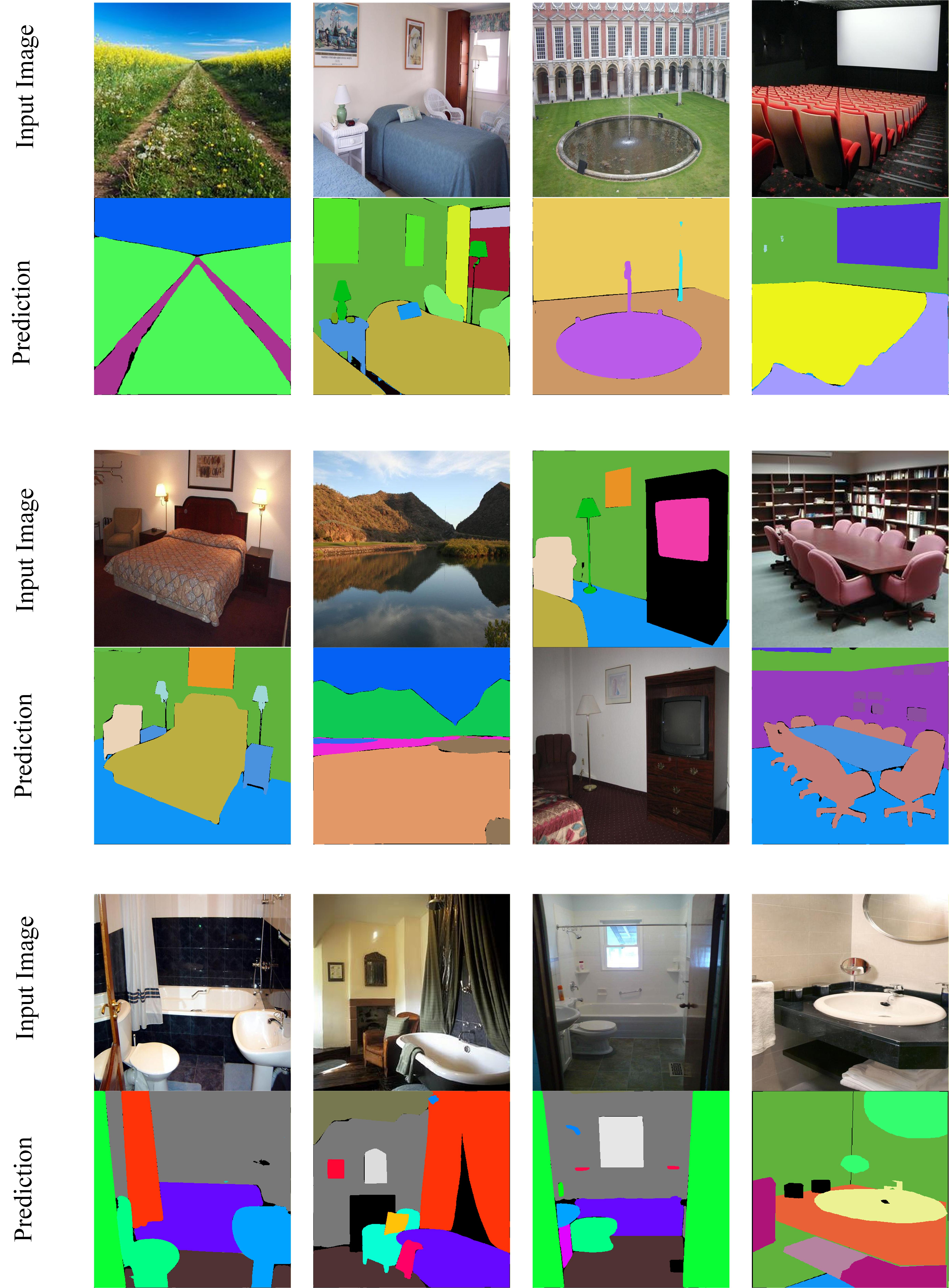}
    \caption{\textbf{Qualitative results} ADE20K on semantic segmentation.}
    \label{visual-5}
\end{figure*}

\begin{figure*}[!t]
    \centering
    \includegraphics[width=0.90\linewidth]{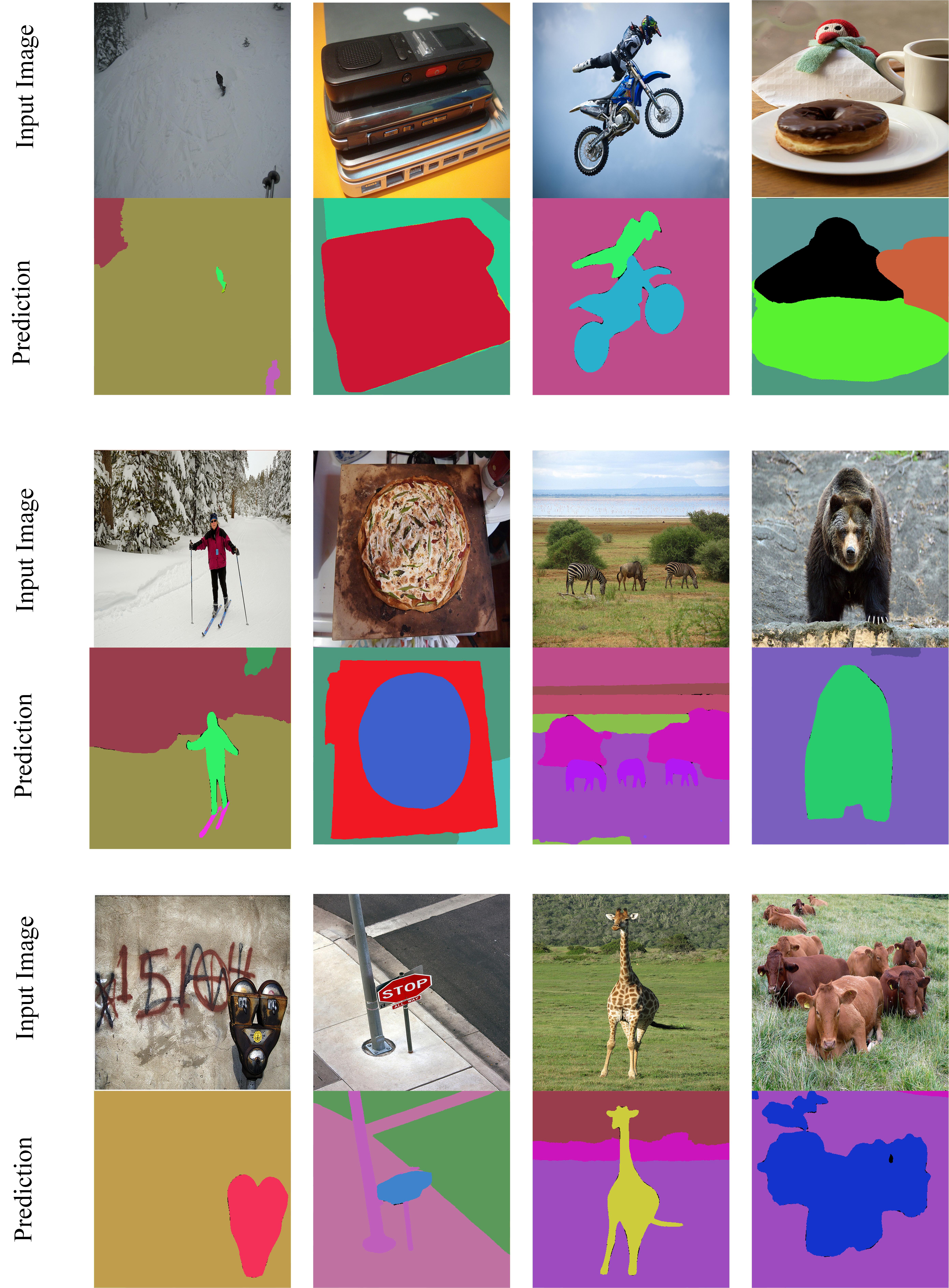}
    \caption{\textbf{Qualitative results} COCO on semantic segmentation.}
    \label{visual-6}
\end{figure*}


%
%
\bibliographystyle{splncs04}
\bibliography{main}
\end{document}